\documentclass[10pt]{article}

\usepackage[preprint]{tmlr}

\usepackage{multirow}
\usepackage{float}
\usepackage{colortbl} 
\usepackage{tcolorbox}   
\usepackage{algorithm}
\usepackage{algpseudocode}
\usepackage{tikz}
\usepackage{color, colortbl}
\usepackage{amsmath}
\usepackage[utf8]{inputenc} 
\usepackage[T1]{fontenc}    
\usepackage{booktabs}       
\usepackage{amsfonts}       
\usepackage{nicefrac}       
\usepackage{microtype}      
\usepackage{float}
\usepackage{wrapfig}
\usepackage{graphicx} 
\usepackage{enumitem}
\usepackage{mdframed}
\usepackage{subcaption}
\usepackage[table]{xcolor}            
\tcbuselibrary{listingsutf8}
\usepackage[utf8]{inputenc}
\usepackage{float}

\newcommand{\rowgray}{\rowcolor[gray]{0.9}}
\newcommand{\rowpink}{\rowcolor[rgb]{1.0,0.9,0.9}}
\newcommand{\rowgreen}{\rowcolor[rgb]{0.9,1.0,0.9}}
\newcommand{\rowblue}{\rowcolor[rgb]{0.9,0.95,1.0}}
\newcommand{\rowour}{\rowcolor[rgb]{1.0,1.0,0.85}} 
\newcommand{\gain}[1]{\textcolor{green!55!black}{+#1}}
\newcommand{\loss}[1]{\textcolor{red!70!black}{#1}}

\definecolor{lightblue}{RGB}{235,245,255}

\newmdenv[
  backgroundcolor=lightblue,
  linecolor=blue!50,
  linewidth=0pt,
  innertopmargin=10pt,
  innerbottommargin=10pt,
  innerleftmargin=10pt,
  innerrightmargin=10pt,
  skipabove=10pt,
  skipbelow=10pt,
  roundcorner=4pt
]{blueblock}

\usepackage{graphicx}
\usepackage{booktabs}
\usepackage{url}
\usepackage{hyperref}
\usepackage{cleveref}

\def\month{MM}
\def\year{YYYY}
\def\openreview{\url{https://openreview.net/forum?id=XXXX}}

\title{
SpatialThinker: Reinforcing Scene Graph-Grounded Spatial Reasoning via Dense Rewards
}

{\centering
\author{
\textbf{Hunar Batra}$^{1}$\thanks{Correspondence to \texttt{\{hunar.batra, ronald.clark\}@cs.ox.ac.uk}}\quad
\textbf{Haoqin Tu}$^{2}$\quad
\textbf{Hardy Chen}$^{2}$\quad
\textbf{Yuanze Lin}$^{1}$\quad
\textbf{Cihang Xie}$^{2}$\quad
\textbf{Ronald Clark}$^{1}$\\[6pt]
\hspace{6.5em}$^{1}$University of Oxford \quad $^{2}$University of California, Santa Cruz
}}

\begin{document}

\thispagestyle{fancy}
\pagestyle{fancy}

\maketitle

\vspace{-2.25em}
\begin{center}
{\small
\raisebox{-0.2em}{\includegraphics[height=1.1em]{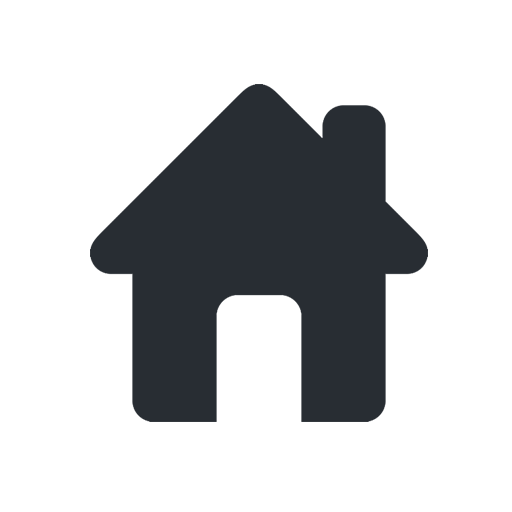}}~
\href{https://pixl.cs.ox.ac.uk/spatial-thinker}{Project page}\hspace{2em}
\raisebox{-0.2em}
{\includegraphics[height=1.1em]{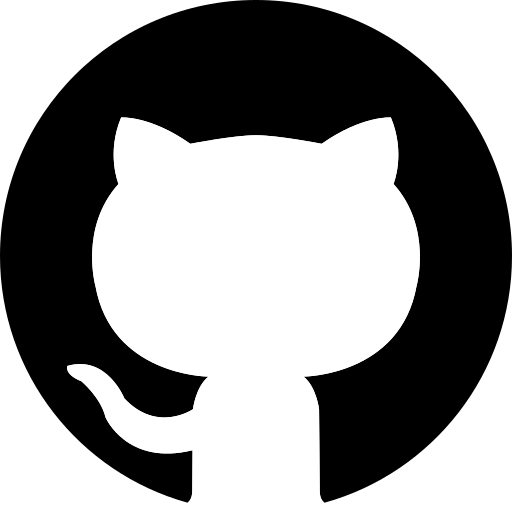}}~
\href{https://github.com/hunarbatra/SpatialThinker}{Code}\hspace{2em}
\raisebox{-0.2em}{\includegraphics[height=1.1em]{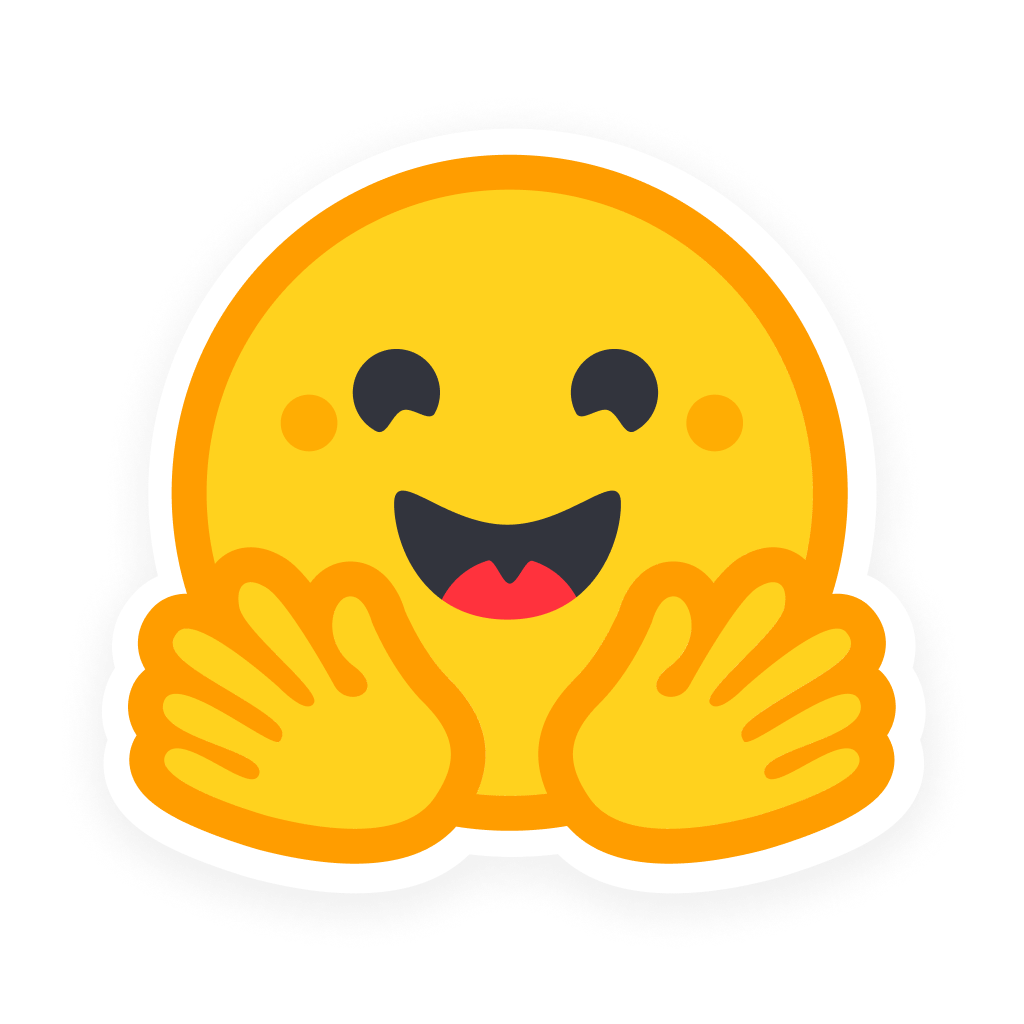}}~
\href{https://huggingface.co/collections/OX-PIXL/spatialthinker}{Models \& Dataset}
}
\end{center}
\vspace{0.5em}

% ...
\begin{figure}[H]
    \centering
    \includegraphics[width=1\linewidth]{images/updated_intro_img.pdf}
    \vspace{-2em}
    \caption{Current MLLMs struggle with fine-grained spatial reasoning, inferring affordances, or interpreting real-world spatial relationships. \textsc{SpatialThinker} addresses this by integrating scene graph generation within reasoning, enforcing structured grounding.}
    \label{fig:main_examples}
\end{figure}

\begin{abstract}

Multimodal large language models (MLLMs) have achieved remarkable progress in vision–language tasks, but continue to struggle with spatial reasoning. Existing spatial MLLMs rely on large-scale datasets, explicit 3D inputs, architecture-specific modifications, or sparse Reinforcement Learning (RL) methods that provide insufficient guidance for spatially-grounded reasoning. We introduce \textsc{SpatialThinker}. To our knowledge, it is the first MLLM unifying Scene Graph Generation (SGG) and visual reasoning in a single pass via online RL. The model simulates human-like spatial perception by constructing a mental scene graph of task-relevant objects and relations, and reasoning toward an answer via dense spatial rewards. Our contributions are threefold: (1) SGG-grounded reasoning: integrating SGG directly within the reasoning chain rather than as a disjoint preprocessing step; (2) STVQA-7K: a high-quality spatial VQA training dataset via a scalable synthesis pipeline; and (3) a dense spatial reward design that enforces structured grounding during RL and generalizes to improve broad visual perception. \textsc{SpatialThinker-7B} achieves 3.6$\times$ larger gains over SFT and $1.7\times$ better in- and out-of-distribution generalization than sparse RL. Trained on only 7K samples, \textsc{SpatialThinker-7B} matches GPT-5 and outperforms GPT-4o, while \textsc{SpatialThinker-30B} surpasses both GPT-5 and Claude 4 Sonnet on average across 14 spatial and real-world benchmarks, demonstrating that structured spatial grounding with reward-aligned reasoning enables robust spatial understanding with limited data.
\end{abstract}

    \section{Introduction}

    Spatial reasoning is central to human intelligence, enabling us to perceive, localize, and manipulate objects in complex environments, a capability crucial for embodied AI tasks like robotic manipulation \cite{Intelligence202505AV, Gao2023PhysicallyGV, Nasiriany2024PIVOTIV}, navigation \cite{Huang2022VisualLM}, and augmented reality \cite{Konenkov2024VRGPTVL}, where precise spatial awareness underpins real-world deployment \cite{Driess2023PaLMEAE, Team2025GeminiRB}. While MLLMs have advanced rapidly in vision-language tasks \cite{hurst2024gpt, lin2024vila, deitke2025molmo, Bai2025Qwen25VLTR, Du2025KimiVLTR, Liu2023VisualIT, google2025gemini2flash}, they continue to struggle with spatial understanding, especially in 3D space, which requires capturing geometry, structure, and relations beyond 2D projections \cite{chen2024spatialvlm, tong2024eyes, kamath2023s, yang2025thinking, Tong2024Cambrian1AF, Ma20243DSRBenchAC}.

    Existing approaches address this through large-scale data synthesis from 3D scene graphs \cite{chen2024spatialvlm, ma2025spatialllm, daxberger2025mm, cheng2024spatialrgpt}, auxiliary spatial tokens or architectural modifications \cite{hong20233d, ma2025spatialllm}, or explicit 3D inputs like depth maps and point clouds \cite{hong2023llm, cheng2024spatialrgpt, cai2024spatialbot}. A common pattern is the use of scene graphs as an offline preprocessing tool for data curation: SpatialRGPT builds 3D scene graphs from point clouds to generate 700K training samples \cite{cheng2024spatialrgpt}, while SpatialVLM and SpatialLLM rely on scene graph-derived annotations to supervise training on 2B and 1M samples respectively \cite{chen2024spatialvlm, ma2025spatialllm}, resulting in data-intensive pipelines that require either massive scale or architecture-specific modifications.

    \begin{figure*}[t]
        \centering
        \includegraphics[width=1\linewidth]{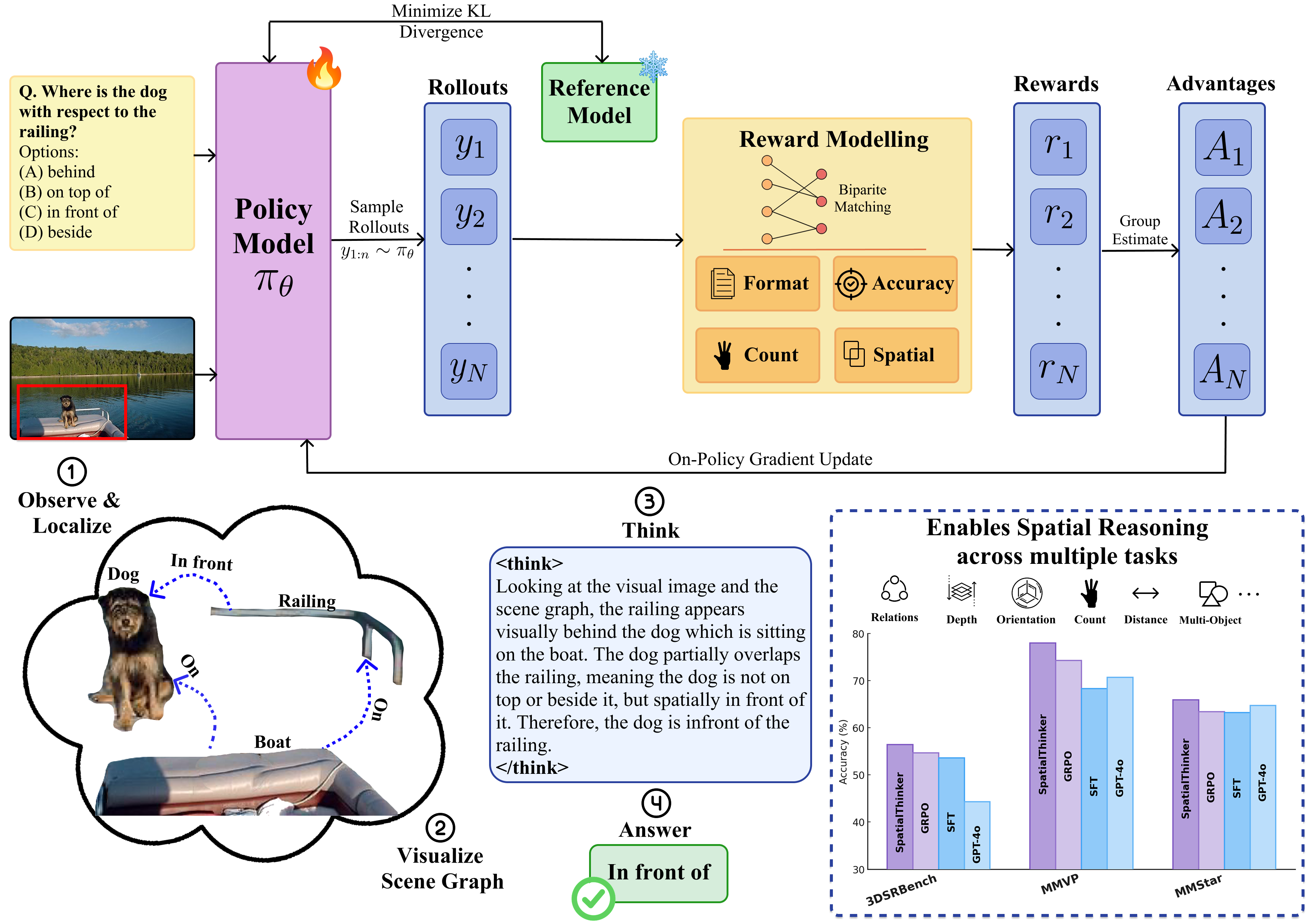}
        \caption{Method overview of \textsc{SpatialThinker}. Our framework unifies scene graph generation and visual reasoning via multi-objective dense RL training, enabling structured spatial grounding as part of the reasoning chain leading to enhanced spatial understanding in MLLMs across tasks and general visual perception.}
        \label{fig:method_overview}
    \end{figure*}

    Recently, RLVR has shown superior generalization over SFT by learning diverse reasoning strategies rather than static patterns \cite{DeepSeekAI2025DeepSeekR1IR, Shen2025VLMR1AS, Gandhi2025CognitiveBT}. However, existing RLVR approaches for spatial reasoning employ simple accuracy-only rewards, providing insufficient guidance for visually-grounded reasoning \cite{Shen2025SATORIR1IM, Xiao2025AdvancingMR, Ma2025SpatialReasonerTE, Wang2025SVQAR1RS, Xia2025VisionaryR1MS, Zhu2025Struct2DAP}. Meanwhile, scene graphs offer natural structure to guide visual reasoning \cite{Hildebrandt2020SceneGR, Wald2020Learning3S}, yet existing methods treat them either as external preprocessing pipelines for data curation \cite{kim2024llm4sgg, chen2023gpt4sgg, li2024pixels} or as isolated generation targets decoupled from downstream reasoning \cite{chen2025compile, li2025relation}, never integrating them end-to-end within the reasoning process itself.

We address these with \textsc{SpatialThinker}, the first MLLM to our knowledge that unifies Scene Graph Generation (SGG) and visual reasoning via online RL. Rather than treating SGG as a disjoint preprocessing step, \textsc{SpatialThinker} integrates scene graph construction directly within the reasoning chain, constructing region-of-interest scene graphs that capture task-relevant objects, spatial relations, and localized coordinates, and reasons over these structured representations toward an answer. The training leverages a multi-objective reward with lexicographic ordering: format rewards enforce structured reasoning; count penalties regulate regional focus; accuracy rewards prioritize correctness; and spatial rewards encourage precise localization. This promotes human-like reasoning following \textit{observe, localize, think, answer}, where 2D grounding tells the model "where objects are" while 3D relational predicates tell it "how objects sit in the world", mirroring the mental sketches humans form when perceiving a scene.

    \textsc{SpatialThinker-7B}, trained on only 7K samples from our synthesized \textsc{STVQA-7K} dataset, outperforms SFT (+5.5\%) and conventional RL baselines (+3.2\%) across fourteen benchmarks, surpassing GPT-4o (+4.7\% avg.), Claude 3.5 Sonnet (+9.6\% avg.), and Claude 4 Sonnet (+1.8\% avg.) \cite{hurst2024gpt, anthropic2024claude35addendum, anthropic2025systemcard} while matching GPT-5 ($-0.9\%$ avg.). While vanilla sparse RL improves the base model by +4.4\%, our dense spatial rewards achieve +7.7\%, nearly doubling ($\times$1.7) the RL benefit. Crucially, sparse RL merely matches SFT on OOD real-world VQA (+2.7\% vs. +2.9\%), whereas \textsc{SpatialThinker} achieves +5.2\%, confirming that accuracy-only rewards fail to improve generalization to generic visual perception tasks. Scaling the same recipe to 30B, \textsc{SpatialThinker-30B} attains the best overall average, surpassing GPT-5 by +3.0\% and Claude 4 Sonnet by +5.8\%, and reaching 93.6\% on CV-Bench 3D \cite{Tong2024Cambrian1AF} ($+3.3\%$ over GPT-5). This demonstrates that structured spatial grounding via dense rewards surpasses static SFT patterns learned from far larger datasets \cite{chen2024spatialvlm, ma2024spatialpin}.

    Our main contributions are:
    \begin{itemize}
        \item We propose \textsc{SpatialThinker}, the first MLLM to our knowledge that unifies SGG and visual reasoning generation in a single pass via online RL, enabling the model to jointly perceive spatial structure and reason, achieving strong performance with only 7K training samples vs. hundreds of thousands to billions used by existing Spatial MLLMs. 
        
        \item We introduce \textsc{STVQA-7K}, a high-quality spatial VQA training dataset grounded in scene graphs annotations, along with a scalable data generation pipeline supporting up to 108K samples with dual-LLM verification for quality control.
        
        \item We design a dense, lexicographically gated multi-objective reward that guides regionally focused spatial reasoning, and generalizes beyond spatial tasks to improve broad visual perception, achieving superior in- and out-of-distribution generalization across spatial, generic VQA, and real-world benchmarks, and outperforming conventional RL and SFT baselines, open-sourced generalist and spatial MLLMs, and proprietary models. We release \textsc{SpatialThinker} at 3B, 7B, and 30B scales, with \textsc{SpatialThinker-30B} surpassing state-of-the-art proprietary models including GPT-5 and Claude 4 Sonnet on average across 14 benchmarks.
        
    \end{itemize}

\section{Preliminaries}

\textbf{Scene Graph Generation (SGG).}
A scene graph provides a structured representation of an image \(I\) as a directed graph \(G=(V,E)\). 
Each node \(v_i \in V\) denotes an object with a category label \(c_i\) and a 2D bounding box 
\(b_i=(x_1,y_1,x_2,y_2)\); each edge \(e_{ij}\in E\) is a relationship triplet 
\(\langle v_i, r_{ij}, v_j \rangle\) consisting of subject $v_i$, predicate $r_{ij}$, and object $v_j$ that capture spatial or interactive relations (e.g., \emph{left of}, \emph{on}, \emph{under}) \citep{Hildebrandt2020SceneGR, Wald2020Learning3S}. Classical SGG decomposes prediction into object detection and relation recognition \citep{carion2020end, cong2023reltr}, while open-vocabulary methods leverage language or vision priors to generalize beyond fixed ontologies \citep{Chen2024SceneGG, Li2023ZeroshotVR}. 
We refer to \emph{question-focused scene subgraphs} as \(G_q=(V_q,E_q)\subseteq G\) that retain objects and relations relevant to a given query \(q\).

\textbf{Reasoning in Multimodal Large Language Models.}  
MLLMs aim to solve reasoning tasks defined over a dataset $\mathcal{D}$ of multimodal instances $(\mathbf{x}_{\text{img}}, \mathbf{x}_{\text{text}}, \mathbf{y}^*)$, where $\mathbf{x}_{\text{img}}$ is a visual input, $\mathbf{x}_{\text{text}}$ is a natural language query, and $\mathbf{y}^*$ is the ground-truth answer. We model the MLLM as an autoregressive policy $\pi_\theta$ that outputs a trajectory $\mathbf{y} = (s_1, \dots, s_T, a)$ consisting of reasoning steps $s_t$ and a final answer $a$. The policy factorizes as:
\begin{equation}
\pi_\theta(\mathbf{y} \mid \mathbf{x}_{\text{img}}, \mathbf{x}_{\text{text}}) = \left( \prod_{t=1}^{T} \pi_\theta(s_t \mid \mathbf{x}_{\text{img}}, \mathbf{x}_{\text{text}}, s_{<t}) \right) \cdot \pi_\theta(a \mid \mathbf{x}_{\text{img}}, \mathbf{x}_{\text{text}}, s_{\leq T}) \
\end{equation}

Supervised fine-tuning enables imitation of reference reasoning traces but often struggles with out-of-distribution generalization. Reinforcement learning (RL) instead optimizes reasoning trajectories with explicit reward signals, improving robustness  \citep{Gandhi2025CognitiveBT, DeepSeekAI2025DeepSeekR1IR, Huang2025VisionR1IR}. The RL objective is given by: $\max_\theta \; \mathbb{E}_{(\mathbf{x}_{\text{img}}, \mathbf{x}_{\text{text}}, \mathbf{y}^*) \sim \mathcal{D}, \; \mathbf{y} \sim \pi_\theta} \big[ R(\mathbf{y}) \big],$ where $R(\mathbf{y})$ evaluates the trajectory based on format adherence, object counting, answer correctness, and spatial localization.

\section{SpatialThinker: Scene Graph-Grounded Spatial Reasoning MLLMs}

\paragraph{Task Formulation}
We cast spatial reasoning in MLLMs as the task of producing a visually grounded response $\mathbf{y}$ to a query $Q = \{\mathbf{x}_{\text{img}}, \mathbf{x}_{\text{text}}\}$. Unlike generic reasoning, our formulation explicitly requires constructing question-focused scene subgraphs $G_q$ and reasoning over objects, bounding boxes, and relations. The policy $\pi_\theta$ is trained on spatially grounded VQA samples from our STVQA-7K dataset (\ref{sec:data}) using our multi-objective spatial reward design $R$ (\ref{sec:reward}) verified based on the ground-truth scene graphs from the dataset. This enforces structural validity, count fidelity, answer accuracy, and precise spatial grounding.


\subsection{Multi-Objective Reward Design}\label{sec:reward}
\textsc{SpatialThinker} is trained with a fine-grained, multi-objective reward function that guides spatial reasoning via explicit visual grounding. Unlike prior RLVR methods that use sparse final-answer rewards \citep{Peng2025LMMR1E3, Zhu2025Struct2DAP, Shen2025VLMR1AS}, our dense reward design combines lexicographic gating with four components—format, count, accuracy, and spatial rewards. We further discuss our reward design process, including ablations and our rationale in Appendix \ref{sec:reward_rationale}.


\textbf{Format Reward.} We enforce a visually-grounded and structured reasoning template: \texttt{<observe>} for scene description, \texttt{<scene>} for regional scene graphs with objects, bounding boxes, and relations, \texttt{<think>} for explicit reasoning, and \texttt{<answer>} for the final output. Beyond tag presence, the format reward validates the JSON inside \texttt{<scene>}, ensuring (1) it is parseable, (2) each object includes required fields (ID and bounding box), and (3) all relations are valid subject–predicate–object triplets. This encourages sequential grounding: perceive → localize → reason → answer. The reward $R_f \in {0,1}$ is weighted at $w_{\text{format}} = 0.1$.


\textbf{Accuracy Reward.} To prioritize task performance, we define the accuracy reward $R_a$ as a binary score based on exact string match between the model’s predicted answer and the ground-truth answer, enabled by our multiple-choice format. This component carries the highest weight ($w_{\text{accuracy}} = 0.5$), directly incentivizing correct final predictions, while the other rewards shape how the model arrives at correct answers.

\textbf{Count Reward.} The count reward encourages the model to predict the appropriate number of objects and relations relevant to the query, penalizing both under- and over-generation based on the deviation between predicted and ground-truth counts for both:

\begin{center}
{\small
\scalebox{0.85}{$
\begin{aligned}
R_{\text{count}} 
= w_{\text{count}} \cdot \Bigg(
    \lambda_{\text{obj}} \cdot 
    \max\!\left(0,\; 1 - 
        \frac{|N_{\text{obj}}^{\text{pred}} - N_{\text{obj}}^{\text{gt}}|}
             {\max(N_{\text{obj}}^{\text{gt}}, 1)}
    \right) + \lambda_{\text{rel}} \cdot 
    \max\!\left(0,\; 1 - 
        \frac{|N_{\text{rel}}^{\text{pred}} - N_{\text{rel}}^{\text{gt}}|}
             {\max(N_{\text{rel}}^{\text{gt}}, 1)}
    \right)
\Bigg)
\end{aligned}
$}
}
\end{center}

where $N^{\text{pred}}$ and $N^{\text{gt}}$ denote predicted and ground truth counts respectively, $w_{\text{count}} = 0.2$ is the overall count reward weight, and $\lambda_{obj}$ and $\lambda_{rel}$ are set to 0.7 and 0.3 respectively. This guides the model to stay focused on question-relevant regions. Without it, we found the models tend to game the spatial reward by generating excessive objects and relations to maximize random matches—a form of reward hacking.



\textbf{Spatial Reward.} To supervise object localization, we compute the spatial reward only when the final answer is correct. Predicted and ground-truth objects are matched using the Hungarian algorithm for bipartite matching with a cost function that combines Complete IoU (CIoU) and semantic similarity:  
{\small
\begin{equation}
C(o_i^{\text{pred}}, o_j^{\text{gt}}) 
= \lambda_{\text{spatial}} \left(1 - \text{IoU}(b_i, b_j)\right)
+ \lambda_{\text{semantic}} \left(1 - \text{sim}(l_i, l_j)\right) \
\end{equation}
}

where \(b\) and \(l\) denote bounding boxes and labels, respectively, $\lambda_{spatial}=1.0$, and $\lambda_{semantic}=2.0$. The reward is then computed as the average CIoU across matched pairs:  
\(R_{\text{spatial}} = w_{\text{spatial}} \cdot \left(\frac{1}{|\mathcal{M}|} \sum_{(i,j) \in \mathcal{M}} \text{CIoU}(b_i^{\text{pred}}, b_j^{\text{gt}})\right)\), where $w_{\text{spatial}} = 0.2$. CIoU offers dense supervision over IoU, even for non-overlapping boxes by incorporating distance and aspect ratio terms \citep{Zheng2020EnhancingGF}. Although computed in 2D, these grounding signals combined with 3D-linked relations promote 3D-consistent spatial reasoning.


\textbf{Lexicographic Gating.}
We apply lexicographic ordering with conditional gating \citep{Skalse2022LexicographicMR}, prioritizing format $\succ$ \{count, accuracy\} $\succ$ spatial. The model must first satisfy formatting, then jointly optimize count and accuracy, and receives spatial reward only when the answer is correct. This ensures spatial grounding reinforces valid reasoning and makes the model robust to imperfect intermediate scene graphs. Without accuracy gating, we find that models tend to over-optimize intermediate spatial rewards at the expense of final answer correctness. The final reward is computed as follows, where \(\mathbb{I}[\cdot]\) is the indicator function:\\
\\
\vspace{-1em}  
{\small
\begin{equation}
R_{\text{total}} =
\mathbb{I}[R_{\text{format}} = 1] \cdot \Big(
w_{\text{format}} R_f +
w_{\text{count}} R_c 
\\
+ w_{\text{accuracy}} R_a
\mathbb{I}[R_{\text{accuracy}} = 1]\, w_{\text{spatial}} R_s
\Big)
\end{equation}
}

\subsection{Online RL Policy Optimization}\label{sec:rl}

To train \textsc{SpatialThinker} with dense, lexicographically gated rewards, we adopt Group-Relative Policy Optimization (GRPO) \citep{DeepSeekAI2025DeepSeekR1IR, Shao2024DeepSeekMathPT}, an online RL method that avoids critic networks by estimating advantages through intra-group comparisons. Given an input \(\mathbf{x}\), we sample \(N\) trajectories \(\{y^{(1)}, \ldots, y^{(N)}\}\) from the current policy \(\pi_{\theta_{\text{old}}}\). Each response is scored via our dense spatial reward function (\ref{sec:reward}), and advantages are computed using group-normalized scores:  
\(
A^{(i)} = \frac{r^{(i)} - \mu}{\sigma + \varepsilon}
\),  
where \(\mu\) and \(\sigma\) are the group mean and standard deviation, and \(\varepsilon = 10^{-6}\).
We then update the policy using a PPO-style clipped loss with KL regularization:
\begin{equation}
\mathcal{L}_{\text{RL}}(\theta)
= - \frac{1}{G} \sum_{i=1}^G \frac{1}{|y^{(i)}|}
\sum_{t=1}^{|y^{(i)}|}
\Big[
\min\!\big( r^{i,t} A^{(i)},\;
\text{clip}(r^{i,t},\, 1 - \epsilon_l,\, 1 + \epsilon_h)\, A^{(i)} \big)
- \beta \, D^{i,t}_{\text{KL}}
\Big]
\end{equation}

where $r^{i,t} = \frac{\pi_\theta(y^{(i)}_t \mid \mathbf{x}, y^{(i)}_{<t})}{\pi_{\theta_{\text{old}}}(y^{(i)}_t \mid \mathbf{x}, y^{(i)}_{<t})}$ is the importance ratio between new and old policies, and \(D^{i,t}_{\text{KL}}\) is the token-level KL divergence against a reference model. We set \(\epsilon_l = 0.2\), \(\epsilon_h = 0.3\), and \(\beta = 10^{-2}\). This objective balances learning from dense spatial rewards while constraining policy divergence to ensure stability and generalization.

\subsection{STVQA-7K: Dataset Construction}\label{sec:data}


To facilitate reward-aligned spatial reasoning, we construct STVQA-7K, a synthetic visual question answering (VQA) dataset built from human-annotated scene graphs in Visual Genome \citep{krishna2017visual}. STVQA-7K comprises 7,587 spatially grounded multiple-choice VQA pairs spanning both 2D and 3D spatial understanding, covering nine core reasoning types including relations, size, orientation, distance, depth, reach, location, count, and existence. 

We augment the original VG150 predicate set with 34 additional spatial relations—covering distance (e.g., near, far), size (e.g., bigger, taller), orientation (e.g., facing away), and containment (e.g., inside, beneath)—to enrich the relational vocabulary beyond the standard 50 predicates. From these scene graphs, we generate 3D-reasoning VQA, e.g., depth: ``\textit{Which is closer to the camera?}'' and orientation: ``\textit{From the person's perspective, which direction is the dog?}''. 

\begin{wrapfigure}{r}{0.45\linewidth}
    \centering
    \includegraphics[width=\linewidth]{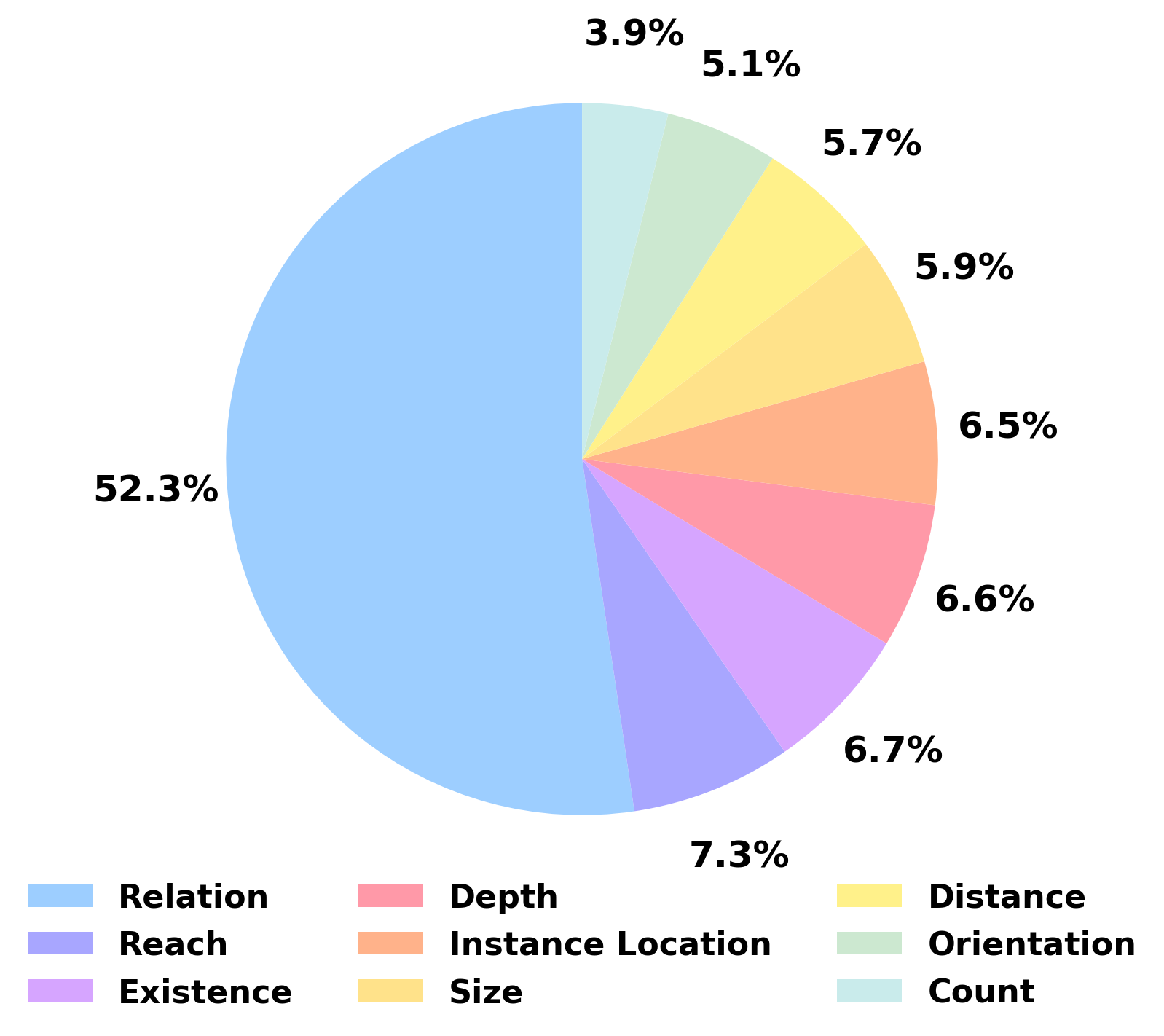}
    \caption{Distribution of STVQA-7K QA types. The dataset covers diverse spatial reasoning aspects, including spatial relations, localization, existence, reach, depth, distance, size, count \& orientation.}
    \label{fig:data_dist}
\end{wrapfigure}

Each QA pair is generated from a scene graph using Claude Sonnet 4 \citep{anthropic2025systemcard}, and rated by difficulty and quality. To mitigate potential biases from LLM-generated questions, we apply a consistency-based dual-LLM filtering pipeline for quality control: Claude Sonnet 4 generates QA pairs, then GPT-4o \citep{hurst2024gpt} validates them via pass@2 agreement. This cross-model verification yields a +13\% accuracy improvement (\autoref{tab:reward_ablation1} , confirming that dual-LLM validation retains high-quality samples. From an initial pool of 56,224 questions, we retain the top 7,587 samples based on rating, difficulty, and verification.

To enable region-specific reasoning, we extract relevant objects and relations per question via lemmatized keyword matching, constructing question-aligned scene subgraphs as localized supervision. This localized supervision helps the model learn where to focus within complex scenes. Bounding box coordinates are retained in absolute pixel space to preserve real-world scale for CIoU-based reward training.

Importantly, our pipeline is scalable and can be extended to generate up to $\sim$108K samples, the maximum supported by Visual Genome, enabling future large-scale post-training or RL fine-tuning. \ref{fig:data_dist} shows the distribution of QA categories. Full dataset construction details and examples are provided in Appendix \ref{sec:full_data}.



\begin{figure*}[t]
    \centering
    \includegraphics[width=1\linewidth]{images/updated_eg_img.pdf}
    \caption{
        \textbf{Qualitative comparison between GPT-5-0807 and SpatialThinker-7B.}
        GPT-5-0807 often fails to distinguish objects at a \textit{3D relational level}—for example, confusing spatial relations such as \textit{beside}, \textit{behind}, and \textit{in front of}, or missing fine-grained object details.
    }
    \label{fig:st-output-eg}
\end{figure*}

\subsection{Training Details}\label{sec:imp_details}
We build \textsc{SpatialThinker} upon three strong open-source multimodal base models: Qwen2.5-VL-3B, Qwen2.5-VL-7B \citep{Bai2025Qwen25VLTR}, and Qwen3-VL-30B \citep{bai2025qwen3vltechnicalreport}, using them as backbones for policy optimization with RL. No SFT is performed prior to RL training on our STVQA-7K dataset (\ref{sec:data}). We employ GRPO \citep{Shao2024DeepSeekMathPT} as the advantage estimator as described in \ref{sec:rl}, using a rollout size of $8$ samples per query and a sampling temperature of $1.0$. The models are trained with a maximum context length of 16,384 tokens. The rollout batch size is set to 512, and the global batch size is 128. We train for 75 training steps i.e., ~5 training episodes) on 4 $\times$ NVIDIA H100 80GB GPUs. Training time totals $\sim{13}$ hours for the 3B model and $\sim{15}$ hours for the 7B model. To assess scaling, we further apply the same dense-reward GRPO objective to a larger Qwen3-VL-30B-A3B-Instruct backbone \cite{bai2025qwen3vltechnicalreport}, trained with LoRA (rank 64) via the Tinker API, yielding \textsc{SpatialThinker-30B} (details in Appendix~\ref{sec:add_exp_details}).

The models are trained on high-resolution image inputs ranging from $512\times512$ to $2048\times2048$ pixels, to preserve fine-grained spatial information. All model parameters, including the vision encoder, are updated during training. We use the AdamW optimizer with $\texttt{bf16}$ precision, a learning rate of $1 \times 10^{-6}$, and  weight decay of $1 \times 10^{-2}$. The KL penalty coefficient is set to $10^{-2}$ (Ablation in Appendix \ref{sec:kl_abl}). STVQA-7K is partitioned with a 90/10 train–validation split. Further details on prompts, SFT, and RL training setups, are provided in Appendices ~\ref{sec:prompt}, \ref{sec:sft_train}, and \ref{sec:rl_train}, respectively. 
Finally, Appendix \ref{sec:rl_curves} illustrates how each reward component improves steadily under our multi-objective spatial reward, reflecting stable learning dynamics.

\textbf{Inference Overhead.} As \textsc{SpatialThinker} generates question-focused scene \textit{sub}graphs rather than exhaustive scene descriptions, enforced by the count penalty and RoI-filtered supervision (\ref{sec:reward}), the structured reasoning adds only modest overhead: on average ${\sim}$120 additional tokens for the scene graph component.

\section{Experiments}

\begin{table*}[t]
\caption{Performance over 2D \& 3D Spatial Understanding Benchmarks across different model types. Top-1 \& Top-2 accuracies are represented using \textbf{bold text}, and \underline{underlines}.}
\centering
\setlength{\tabcolsep}{20pt} 
\resizebox{0.88\textwidth}{!}{%
\begin{tabular}{l|c|cc|c|cc|c}
\toprule
\multirow{2}{*}{\textbf{Model}} & 
\multirow{2}{*}{\textbf{3DSRBench}} &
\multicolumn{2}{c|}{\textbf{CV-Bench}} & \multirow{2}{*}{\textbf{Avg.}} &
\multicolumn{2}{c|}{\textbf{BLINK$_{val}$}} & \multirow{2}{*}{\textbf{Avg.}} \\
 & & 2D & 3D & & \shortstack{Spatial\\Relation} & \shortstack{Relative\\Depth} & \\
\midrule
\rowgray \multicolumn{8}{c}{\textit{Proprietary Models}} \\
GPT-5-0807 & \textbf{68.2} & \textbf{81.4} & \underline{90.3} & \underline{85.8} & \textbf{90.9} & \textbf{81.4} & \textbf{86.1} \\
GPT-4o-0513 & 44.3 & 75.8 & 83.0 & 79.4 & 82.5 & 78.2 & 80.4 \\
Claude-4-Sonnet-0514 & 61.9 & 73.3 & 84.2 & 78.7 & 79.0 & 78.2 & 78.6 \\
Claude-3.5-Sonnet-0620 & 48.2 & 60.2 & 71.5 & 65.9 & 58.7 & 67.7 & 63.2 \\
\midrule
\rowpink \multicolumn{8}{c}{\textit{Open-Source General MLLMs}} \\
Qwen2.5-VL-3B & 44.0 & 59.9 & 60.2 & 60.0 & 66.4 & 54.0 & 60.2 \\
Qwen2.5-VL-7B & 48.4 & 69.1 & 68.0 & 68.6 & 84.0 & 52.4 & 68.2 \\
Qwen3-VL-30B & 60.4 & 79.0 & 89.6 & 84.3 & 86.0 & 75.8 & 80.9 \\
VLAA-Thinker-Qwen2.5-VL-7B & 52.2 & 60.8 & 60.3 & 60.6 & 81.2 & 71.0 & 76.1 \\
LLaVA-NeXT-8B & 48.4 & 62.2 & 65.3 & 63.8 & - & - & - \\
Cambrian-1-8B & 42.2 & 72.3 & 72.0 & 72.2 & 69.9 & 73.4 & 71.7 \\
\midrule
\rowgreen \multicolumn{8}{c}{\textit{Open-Source Spatial MLLMs}} \\
RoboPoint-13B & - & - & 61.2 & - & 60.8 & 61.3 & 61.1 \\
SpatialBot-3B & 41.1 & - & 69.1 & - & 67.8 & 67.7 & 67.8 \\
SpaceLLaVA-13B & 42.0 & - & 68.5 & - & 72.7 & 62.9 & 67.8 \\
SATORI-R1 & 48.0 & 54.6 & 69.4 & 62.0 & 77.0 & 58.9 & 68.0 \\
Spatial-RGPT-7B w/ depth & 48.4 & - & 60.7 & - & 65.7 & 72.3 & 69.0 \\
SpaceThinker & 51.1 & 65.1 & 65.9 & 65.5 & 73.4 & 59.9 & 66.7 \\
SpaceOm & 52.2 & 72.1 & 69.3 & 70.7 & 81.1 & 65.3 & 73.2 \\
\midrule
\rowblue \multicolumn{8}{c}{\textit{Method Comparison (Trained on STVQA-7K)}} \\
Qwen2.5-VL-3B + SFT & 50.8 & 53.9 & 68.4 & 61.1 & 65.0 & 66.9 & 66.0 \\
Qwen2.5-VL-3B + Vanilla GRPO & 50.1 & 70.6 & 66.6 & 68.6 & 73.4 & 55.6 & 64.5 \\
\rowour \textbf{SpatialThinker-3B (Ours)} & 52.9 & 71.0 & 76.3 & 73.6 & 81.8 & 66.9 & 74.4 \\
Qwen2.5-VL-7B + SFT & 53.6 & 56.1 & 71.3 & 63.7 & 75.5 & 64.5 & 70.0 \\
Qwen2.5-VL-7B + Vanilla GRPO & 54.7 & 68.9 & 76.5 & 72.7 & 80.4 & 75.0 & 77.7 \\
\rowour \textbf{SpatialThinker-7B (Ours)} & 56.4 & 77.7 & 78.7 & 78.2 & 86.0 & 72.6 & 79.3 \\
\rowour \textbf{SpatialThinker-30B (Ours)} & \underline{62.1} & \underline{80.3} & \textbf{93.6} & \textbf{87.0} & \underline{88.1} & \underline{79.8} & \underline{84.0} \\
\bottomrule
\end{tabular}}
\vspace{-1em}
\label{tab:spatialthinker_results}
\end{table*}


We evaluate \textsc{SpatialThinker} across 14 diverse spatial understanding, and real-world and generic VQA benchmarks, encompassing both 2D and 3D reasoning tasks. Our experiments are guided by two core questions:
(Q1) Does our spatial VQA generation pipeline, combined with dense reward RL, improve general spatial and visual reasoning in MLLMs?
(Q2) Can MLLMs learn strong spatial capabilities from just 7K synthetic training samples, and how does this compare to models trained on orders-of-magnitude more data?

\textbf{Benchmarks.}  
We evaluate across 14 benchmarks: eight spatial and six real-world VQA. The spatial suite comprises eight benchmarks including CV-Bench 2D and 3D \citep{Tong2024Cambrian1AF}, BLINK Spatial Relations and Relative Depth \citep{Fu2024BLINKML}, 3DSRBench \citep{Ma20243DSRBenchAC}, MMVP \citep{tong2024eyes}, SpatialBench \citep{cai2024spatialbot}, and SpatialReasonerEval \citep{Ma2025SpatialReasonerTE}; the multi-view MindCube-tiny \citep{wang2026mindcubespatialmentalmodeling}; and the held-out validation split of our released STVQA-7K test set, which spans nine spatial reasoning categories: spatial relations, reach and interaction, comparative size, orientation, instance location, depth ordering, distance comparison, object counting, and existence. Together these cover spatial relations, depth, distance, counting, size, orientation, and egocentric multi-view 3D reasoning. To test generalization in real-world, embodied, and generalist VQA contexts, we use six further benchmarks: VStarBench \citep{Wu2023VGV}, RealWorldQA \citep{grok2024vision}, MME-RealWorld \citep{Zhang2024MMERealWorldCY}, RoboSpatial-Home \citep{song2025robospatial} (Configuration and Compatibility only), MM-Star \citep{Chen2024AreWO}, and HallusionBench \citep{Guan2023HallusionbenchAA}, collectively spanning fine-grained visual perception, embodied spatial affordances, structured document and chart understanding, real-world domains (autonomous driving, remote sensing, surveillance), general visual reasoning, and hallucination robustness.

\textbf{Baselines.}  
We compare against proprietary MLLMs including GPT-5 (\textsc{gpt-5-0807}) \citep{openai2025gpt5}, GPT-4o (\textsc{gpt-4o-0513}) \citep{hurst2024gpt}, Claude 4 Sonnet (\textsc{claude-4-sonnet-0514}) \citep{anthropic2025systemcard}, and Claude 3.5 Sonnet (\textsc{claude-3.5-sonnet-0620}) \citep{anthropic2024claude35addendum}, open-source generalist models like Qwen2.5-VL \citep{Bai2025Qwen25VLTR}, Qwen3-VL-30B \citep{bai2025qwen3vltechnicalreport}, LLaVA-NeXT \citep{Li2024LLaVANeXTInterleaveTM}, Cambrian-1 \citep{Tong2024Cambrian1AF}, and VLAA-Thinker \citep{Chen2025SFTOR}, and spatially-tuned open-source MLLMs such as SpaceLLaVA \citep{SpaceLLaVA2025, chen2024spatialvlm}, SpatialRGPT \citep{cheng2024spatialrgpt}, RoboPoint \citep{Yuan2024RoboPointAV}, SpaceThinker \citep{SpaceThinker2025}, SpaceOm \citep{SpaceOm2025}, SpatialReasoner \citep{Ma2025SpatialReasonerTE}, SpatialBot \citep{cai2024spatialbot}, Visionary-R1 \citep{Xia2025VisionaryR1MS}, and SATORI-R1 \citep{Shen2025SATORIR1IM}. We evaluate ablations on variants of our model trained with the STVQA-7K dataset: a supervised fine-tuning (SFT) baseline, and a sparse-reward RL baseline that optimizes only format and accuracy rewards, each weighted equally at 0.5, to isolate the effect of our dense spatial reward.

\textbf{Evaluation Setting.}  
All models are evaluated in a zero-shot setting using greedy decoding (\textsc{temp = 0.0}). Models default prompting format is used where applicable (e.g., for VLAA-Thinker, SpaceOm, SpaceThinker). SpatialRGPT is evaluated with depth inputs; all other models use RGB. Accuracy is the primary evaluation metric. Full benchmark descriptions, baseline details, and additional implementation specifics are provided in Appendix \ref{sec:add_exp_details}.

\begin{table*}[t]
\caption{Performance on additional spatial benchmarks. Top-1 \& Top-2 accuracies are represented using \textbf{bold text}, and \underline{underlines}. STVQA-7K\textsubscript{val} is the held-out in-distribution validation split.}
\centering
\setlength{\tabcolsep}{12pt}
\resizebox{0.78\textwidth}{!}{%
\begin{tabular}{l|ccccc}
\toprule
\textbf{Model} & \textbf{MMVP} &
\textbf{SpatialReasonerEval} &
\textbf{SpatialBench} &
\textbf{STVQA-7K\textsubscript{val}} &
\textbf{MindCube-tiny} \\
\midrule
\rowgray \multicolumn{6}{c}{\textit{Proprietary Models}} \\
GPT-5-0807 & 61.7 & \underline{90.5} & 62.6 & 89.7 & 42.5 \\
GPT-4o-0513 & 70.7 & 85.8 & 67.0 & 77.0 & 34.9 \\
Claude-4-Sonnet-0514 & 71.3 & 85.7 & 60.9 & 80.5 & 44.8 \\
Claude-3.5-Sonnet-0620 & 71.3 & 84.1 & 63.2 & - & - \\
\midrule
\rowpink \multicolumn{6}{c}{\textit{Open-Source General \& Spatial MLLMs}} \\
Qwen2.5-VL-3B & 67.0 & 68.0 & 49.9 & 74.3 & 36.4 \\
Qwen2.5-VL-7B & 72.3 & 70.6 & 62.5 & 77.5 & 35.3 \\
Qwen3-VL-30B & 77.9 & 88.2 & \underline{68.1} & 83.7 & 39.0 \\
VLAA-Thinker-7B & 75.3 & 61.2 & 66.2 & 76.2 & 36.7 \\
SpaceThinker & 63.0 & 69.6 & 57.9 & 75.4 & 36.0 \\
SpaceOm & 66.3 & 68.9 & 58.6 & 66.0 & 33.5 \\
SpatialReasoner & 64.0 & 76.4 & 59.2 & 74.0 & 34.6 \\
SATORI-R1 & 67.7 & 70.5 & 60.3 & 51.2 & 36.0 \\
Visionary-R1 & 70.3 & 72.9 & 59.8 & 74.4 & 36.0 \\
\midrule
\rowblue \multicolumn{6}{c}{\textit{Method Comparison (Trained on STVQA-7K)}} \\
Qwen2.5-VL-3B + SFT & 62.7 & 67.5 & 56.3 & 85.6 & 35.9 \\
Qwen2.5-VL-3B + Vanilla GRPO & 68.3 & 69.3 & 56.9 & 86.7 & 38.8 \\
\rowour \textbf{SpatialThinker-3B (Ours)} & 69.0 & 76.5 & 61.5 & 92.5 & 40.5 \\
Qwen2.5-VL-7B + SFT & 68.3 & 70.8 & 63.5 & 84.5 & 42.2 \\
Qwen2.5-VL-7B + Vanilla GRPO & 74.3 & 79.6 & 64.2 & 87.1 & 44.0 \\
\rowour \textbf{SpatialThinker-7B (Ours)} & \underline{78.0} & 82.7 & 66.4 & \underline{92.8} & \underline{45.1} \\
\rowour \textbf{SpatialThinker-30B (Ours)} & \textbf{79.7} & \textbf{92.6} & \textbf{69.5} & \textbf{93.0} & \textbf{45.4} \\
\bottomrule
\end{tabular}}
\label{tab:benchmarks_mmvp_spatialbench_realworldqa}
\vspace{-1em}
\end{table*}

\subsection{Results}
We evaluate SpatialThinker across eight spatial and six real-world VQA benchmarks to assess its effectiveness in learning spatial understanding and real-world VQA from limited training data through dense reward supervision.

\textbf{Performance across Spatial Benchmarks.} We evaluate \textsc{SpatialThinker} across eight spatial reasoning benchmarks including 2D relational understanding, 3D spatial alignment, multi-view reasoning, counting, depth ordering, and distance comparison. As shown in Tables~\ref{tab:spatialthinker_results} and~\ref{tab:benchmarks_mmvp_spatialbench_realworldqa}, \textsc{SpatialThinker} achieves strong and consistent performance across all spatial tasks. \textsc{SpatialThinker-30B} attains the best results on nearly every benchmark, and \textsc{SpatialThinker-7B} remains the strongest small open-source model. On CV-Bench, \textsc{SpatialThinker-7B} attains an average accuracy of 78.2\% across 2D and 3D tasks, approaching GPT-4o's 79.4\% and leading all comparably-sized open models, while \textsc{SpatialThinker-30B} reaches 87.0\%, exceeding GPT-5 (85.86\%) and Claude 4 Sonnet (78.73\%); on the 3D split specifically, \textsc{SpatialThinker-30B} attains a best-in-class 93.6\% (+3.3\% over GPT-5). On 3DSRBench, which requires orientation and multi-object reasoning, \textsc{SpatialThinker-7B} achieves 56.4\% (surpassing GPT-4o by +12.1\%) and \textsc{SpatialThinker-30B} reaches 62.1\%. On BLINK's spatial relation and relative depth tasks, \textsc{SpatialThinker-7B} achieves 86.0\% and 72.6\% (a 79.3\% average, near GPT-4o's 80.4\%), and \textsc{SpatialThinker-30B} reaches an 84.0\% average, outperforming other spatial MLLMs like Spatial-RGPT-7B (69.0\%), which uses depth inputs and 700K training samples.

\begin{table*}[t]
\caption{Performance on VQA and Real-World benchmarks. Top-1 \& Top-2 accuracies are represented using \textbf{bold text}, and \underline{underlines}.}
\centering
\setlength{\tabcolsep}{2pt}
\resizebox{0.88\textwidth}{!}{%
\begin{tabular}{l|cccccc}
\toprule
\textbf{Model} & \textbf{MM-Star} & 
\textbf{VStarBench} &
\textbf{RealWorldQA} & 
\textbf{MME-RealWorld-Lite} & 
\textbf{RoboSpatial-Home} & 
\textbf{HallusionBench} \\
\midrule
\rowpink \multicolumn{7}{c}{\textit{Proprietary and Open-Source MLLMs}} \\
GPT-5-0807 & 58.9 & 73.3 & \textbf{78.7} & \textbf{57.0} & 71.5 & \underline{73.8} \\
GPT-4o-0513 & 64.7 & 66.0 & \underline{75.4} & \underline{51.6} & 68.4 & 55.0 \\
Claude-4-Sonnet-0514 & 64.4 & 60.7 & 64.0 & 46.9 & 69.7 & 71.2 \\
Claude-3.5-Sonnet-0620 & 65.1 & 51.8 & 60.1 & 45.2 & 57.0 & 55.5 \\
Qwen2.5-VL-3B & 55.9 & 74.9 & 58.2 & 41.9 & 58.7 & 46.3 \\
Qwen2.5-VL-7B & 63.9 & 75.9 & 68.4 & 44.1 & 70.6 & 52.9 \\
Qwen3-VL-30B & 64.3 & 81.2 & 64.8 & 45.8 & 53.1 & 61.5 \\
VLAA-Thinker-7B & 63.8 & 58.1 & 66.4 & 44.6 & 68.9 & 68.9 \\
SpaceThinker & 54.5 & 56.5 & 61.6 & - & 52.6 & 65.4 \\
SpaceOm & 57.7 & 56.5 & 53.3 & - & 68.9 & 62.9 \\
\midrule
\rowblue \multicolumn{7}{c}{\textit{Method Comparison (Trained on STVQA-7K)}} \\
Qwen2.5-VL-3B + SFT & 53.9 & 73.3 & 64.8 & 43.0 & 69.8 & 58.9 \\
Qwen2.5-VL-3B + Vanilla GRPO & 56.7 & 74.3 & 64.4 & 46.7 & 64.0 & 59.0 \\
\rowour \textbf{SpatialThinker-3B (Ours)} & 57.6 & 78.0 & 66.3 & 46.5 & 70.6 & 62.5 \\
Qwen2.5-VL-7B + SFT & 63.2 & 78.0 & 65.4 & 47.4 & 72.4 & 66.2 \\
Qwen2.5-VL-7B + Vanilla GRPO & 63.4 & 73.9 & 66.6 & 46.3 & 76.2 & 60.7 \\
\rowour \textbf{SpatialThinker-7B (Ours)} & \underline{65.9} & \underline{81.7} & 69.2 & 48.3 & \underline{76.3} & 66.4 \\
\rowour \textbf{SpatialThinker-30B (Ours)} & \textbf{66.9} & \textbf{85.9} & 74.9 & 49.2 & \textbf{78.1} & \textbf{75.2} \\
\bottomrule
\end{tabular}}
\label{tab:vqa_realworld_datasets}
\end{table*}

Despite being trained on just 7K synthetic samples and using only RGB inputs, \textsc{SpatialThinker-7B} consistently outperforms open-source baselines, including VLAA-Thinker-7B, Cambrian-1-8B, Spatial-RGPT, SpaceLLaVA, and RoboPoint-13B, all of which are trained on orders of magnitude more data. Notably, it exceeds specialized spatial models as well: on CV-Bench 3D, it outperforms SpaceLLaVA-13B (78.7\% vs. 68.5\%), and on BLINK tasks, it surpasses Spatial-RGPT-7B by +10.3\%, and SpatialBot by +11.5\% despite their reliance on depth information.

On the additional spatial benchmarks (Table~\ref{tab:benchmarks_mmvp_spatialbench_realworldqa}), \textsc{SpatialThinker-30B} is the strongest model overall, topping MMVP (79.7\%), SpatialReasonerEval (92.6\%), and SpatialBench (69.5\%) and surpassing both GPT-5 and Claude 4 Sonnet on each. \textsc{SpatialThinker-7B} is the best small model, with its 78.0\% on MMVP exceeding every baseline, including GPT-4o, Claude 4 Sonnet, and GPT-5 (61.7\%). These results highlight the effectiveness of our dense reward design in enabling generalizable spatial reasoning without the need for explicit geometric inputs or large-scale pretraining. Notably, even though training uses only RGB images and 2D rewards, the relational scene-graph supervision encodes cues linked to depth and orientation. The RL objective trains the model to maintain geometric–relational consistency, which results in emergent 3D reasoning capabilities. Qualitative examples of model outputs are shown in \ref{fig:st-output-eg}, with additional comparisons in Appendix \ref{sec:add_model_eg}. SpatialThinker demonstrates stronger 3D spatial grounding and fine-grained object distinction.

\vspace{-0.5em}
\begin{wraptable}{r}{0.5\textwidth}
\caption{Avg. accuracy across 14 benchmarks (eight spatial and six real-world with relative gains ($\Delta$). 
}
\centering
\setlength{\tabcolsep}{3pt}
\resizebox{0.5\textwidth}{!}{%
\begin{tabular}{l|ccccc}
\toprule
\textbf{Model} &
\textbf{Avg. Acc. (14)} &
$\mathbf{\Delta_{\text{Base}}}$ &
$\mathbf{\Delta_{\text{GPT-5}}}$ &
$\mathbf{\Delta_{\text{GPT-4o}}}$ &
$\mathbf{\Delta_{\text{Claude 4}}}$ \\
\midrule
\rowpink \multicolumn{6}{c}{\textit{Proprietary and Base MLLMs}} \\
GPT-5-0807 & \underline{71.5} & - & - & - & - \\
GPT-4o-0513 & 65.8 & - & - & - & - \\
Claude-4-Sonnet-0514 & 68.7 & - & - & - & - \\
Claude-3.5-Sonnet-0620 & 60.9 & - & - & - & - \\
Qwen2.5-VL-3B & 56.8 & - & - & - & - \\
Qwen2.5-VL-7B & 62.8 & - & - & - & - \\
Qwen3-VL-30B & 68.1 & - & - & - & - \\
\midrule
\rowblue \multicolumn{6}{c}{\textit{Method Comparison (Trained on STVQA-7K)}} \\
Qwen2.5-VL-3B + SFT & 60.7 & \gain{3.9} & \loss{-10.8} & \loss{-5.1} & \loss{-8.0} \\
Qwen2.5-VL-3B + Vanilla GRPO & 62.0 & \gain{5.2} & \loss{-9.5} & \loss{-3.8} & \loss{-6.7} \\
\rowour \textbf{SpatialThinker-3B (Ours)} & 65.9 & \gain{9.1} & \loss{-5.6} & \gain{0.1} & \loss{-2.8} \\
Qwen2.5-VL-7B + SFT & 64.9 & \gain{2.1} & \loss{-6.6} & \loss{-0.9} & \loss{-3.8} \\
Qwen2.5-VL-7B + Vanilla GRPO & 67.2 & \gain{4.4} & \loss{-4.3} & \gain{1.4} & \loss{-1.5} \\
\rowour \textbf{SpatialThinker-7B (Ours)} & 70.5 & \gain{7.7} & \loss{-1.0} & \gain{4.7} & \gain{1.8} \\
\rowour \textbf{SpatialThinker-30B (Ours)} & \textbf{74.5} & \textbf{\gain{6.4}} & \textbf{\gain{3.0}} & \textbf{\gain{8.7}} & \textbf{\gain{5.8}} \\
\bottomrule
\end{tabular}}
\label{tab:avg_accuracy_gains}
\end{wraptable}

\textbf{Performance across Real-World and General VQA Benchmarks.}
We further assess generalization to real-world visual question answering using six diverse benchmarks (Table~\ref{tab:vqa_realworld_datasets}). Compared to the base model, \textsc{SpatialThinker-7B} achieves 65.9\% on MM-Star (+2.0\%), 81.7\% on VStarBench (+5.8\% over base, and +15.7\% over GPT-4o), and 76.3\% on RoboSpatial-Home (+5.7\% over base, and +7.9\% over GPT-4o), surpassing all open-source and proprietary baselines. On hallucination-sensitive and real-world benchmarks, it scores 66.4\% on HallusionBench (+13.5\% over base, and 11.4\% over GPT-4o), 69.2\% on RealWorldQA, and 48.3\% on MME-RealWorld-Lite (+4.2\%). These results demonstrate that dense spatial rewards improve broad visual perception and not just spatial understanding, enhancing visual reasoning in the wild. We attribute these gains to spatial grounding forcing the model to attend to specific image regions and their geometric properties, which reduces hallucination by anchoring reasoning to visual evidence (+13.5\% on HallusionBench), sharpens fine-grained visual distinction (+5.8\% on VStarBench), and improves embodied scene understanding (+5.7\% on RoboSpatial-Home). Scaling to 30B extends these gains further: \textsc{SpatialThinker-30B} attains the best result among all models on MM-Star (66.9\%), VStarBench (85.9\%), RoboSpatial-Home (78.1\%), and HallusionBench (75.2\%, exceeding GPT-5's 73.82\%), improving over its Qwen3-VL-30B base by +25.0\% on RoboSpatial-Home and +13.7\% on HallusionBench.

\textbf{RL Training with Dense Rewards Enables Superior Generalization.}
To isolate the contributions of our dense spatial reward design, we compare against two ablation variants: supervised fine-tuning (SFT) and reinforcement learning with sparse rewards (Vanilla GRPO) using only format and answer accuracy, trained on STVQA-7K. As shown in Table~\ref{tab:avg_accuracy_gains}, \textsc{SpatialThinker-7B} achieves an average accuracy of 70.5\% across all 14 benchmarks, exceeding the gains from SFT by +5.5\% and the sparse GRPO variant by +3.2\%, while essentially matching GPT-5 ($-0.9\%$ avg.) and beating GPT-4o, Claude 4 Sonnet, and every open baseline. Scaling the same recipe to 30B, \textsc{SpatialThinker-30B} reaches 74.5\%, surpassing GPT-5 by +3.0\% and Claude 4 Sonnet by +5.8\%. These gains are consistent at 3B, where \textsc{SpatialThinker-3B} outperforms its SFT and GRPO counterparts by +5.2\% and +3.9\% respectively. Notably, Vanilla GRPO provides only modest improvements over the base model (+4.4 for 7B, +5.2 for 3B), whereas our dense spatial reward raises this to +7.7\% and +9.1\% ($\sim$1.7$\times$), underscoring the complementary learning signal provided by count and spatial objectives, along with lexicographic reward gating. Overall, these results affirm that structured reinforcement learning with dense spatial supervision significantly enhances the spatial and generic VQA capabilities of multimodal LLMs, with a small fraction of high-quality data.

\begin{wraptable}{r}{0.5\textwidth}
\caption{Reward ablation on STVQA-7K\textsubscript{val}.}
\centering
\setlength{\tabcolsep}{4pt}
\resizebox{0.5\textwidth}{!}{
\begin{tabular}{l|c}
\toprule
\textbf{Reward Components} & \textbf{STVQA-7K\textsubscript{val}} \\
\midrule
Format + Accuracy & 74.9 \\
\hspace{1em}+ Spatial & 23.7 \\
\hspace{1em}+ Count & 61.7 \\
\hspace{1em}+ Lexicographic Gating \& RoI Filtering & 76.3 \\
\rowcolor{yellow!25}\hspace{1em}+ Filtered Dataset (pass@2) & \textbf{87.9\textsubscript{\textbf{\textcolor{green!55!black}{(+13.0)}}}} \\
\bottomrule
\end{tabular}}
\label{tab:reward_ablation1}
\end{wraptable}

\textbf{Reward Design Ablation.}
To validate our reward formulation, we conduct a controlled ablation study on the STVQA-7K\textsubscript{val} set, progressively introducing each component and constraint as shown in Table \ref{tab:reward_ablation1}. 
Naively adding spatial rewards causes a reward hacking behavior (23.7\%), as models overgenerate cluttered boxes to exploit the CIoU reward. Introducing the count reward mitigates this (+38\% relative gain), regularizing the object and relations count within the scene graph to match the ground-truth quantities. However, rewarding all scene objects biases the model toward exhaustive descriptions. To address this, we shift to local supervision—rewarding only Regions of Interest (RoIs) tied to question-relevant entities—and apply lexicographic gating to ensure spatial rewards are granted only when the final answer is correct, preventing the model from over-optimizing intermediate process rewards at the expense of outcome accuracy. These adjustments recover and slightly exceed baseline performance (76.3\%), while additional dataset filtering via pass@2 correctness verification with GPT-4o to retain only high-quality, validated samples (7K) yields a further boost to 87.9\%. This staged reward shaping process proves essential for stabilizing optimization and grounding learning in verifiable spatial reasoning.
We discuss the full reward design process details in \ref{sec:reward_rationale}.

\begin{wraptable}{r}{0.5\textwidth}
\caption{Average accuracy gains ($\Delta$) over respective base models on all 13 held-out benchmarks (seven spatial and six real-world VQA), i.e., the full suite minus the in-domain STVQA-7K\textsubscript{val} split.}
\centering
\setlength{\tabcolsep}{4pt}
\resizebox{0.5\textwidth}{!}{%
\begin{tabular}{l|c|c}
\toprule
\textbf{Model Variant} & \textbf{Spatial VQA $\boldsymbol{\Delta}_{\text{Base}}$} & \textbf{Real-World VQA $\boldsymbol{\Delta}_{\text{Base}}$} \\
\midrule
Qwen2.5-VL-3B + SFT & \gain{2.1} & \gain{4.6} \\
Qwen2.5-VL-3B + GRPO & \gain{4.4} & \gain{4.9} \\
\rowour \textbf{SpatialThinker-3B} & \textbf{\gain{9.0}} & \textbf{\gain{7.6}} \\
\midrule
Qwen2.5-VL-7B + SFT & \gain{0.9} & \gain{2.8} \\
Qwen2.5-VL-7B + GRPO & \gain{5.9} & \gain{1.9} \\
\rowour \textbf{SpatialThinker-7B} & \textbf{\gain{8.6}} & \textbf{\gain{5.3}} \\
\bottomrule
\end{tabular}
}
\label{tab:ood_generalization}
\end{wraptable}
\textbf{Out-of-Distribution Generalization: Dense Rewards Enable Stronger Transfer.}
While both SFT and sparse-reward GRPO improve spatial reasoning over base models, their ability to generalize to out-of-distribution (OOD) real-world tasks is limited, when compared to \textsc{SpatialThinker} models. As shown in \ref{tab:ood_generalization}, sparse-reward GRPO provides solid spatial gains over its respective base model (+4.4\% for 3B, +5.9\% for 7B), but offers only marginal improvements on real-world benchmarks (+4.9\% and +1.9\% respectively), nearly matching or underperforming SFT (+4.6\% for 3B, +2.8\% for 7B). In contrast, \textsc{SpatialThinker}, trained with dense spatial and count rewards, achieves significantly stronger OOD generalization: +7.6\% for 3B and +5.3\% for 7B, outperforming baselines at both scales, and \textsc{SpatialThinker-7B} delivers nearly $3\times$ the real-world gains of sparse-reward GRPO (+5.3\% vs.\ +1.9\%), highlighting the robustness of our reward design.
The combination of structured reasoning formats and lexicographically gated rewards encourages models to internalize spatial priors and compositional patterns that transfer effectively to OOD tasks. Appendix \ref{sec:abs_reason} further demonstrates generalization to abstract and multi-view reasoning tasks.

\section{Related Work}

\textbf{3D Spatial Reasoning in MLLMs.} While MLLMs have advanced core visual tasks \citep{hurst2024gpt, lin2024vila, deitke2025molmo, Bai2025Qwen25VLTR, Du2025KimiVLTR, Li2024LLaVANeXTInterleaveTM}, their spatial reasoning abilities remain limited \citep{mirzaee2021spartqa, tong2024eyes, kamath2023s, yamada2023evaluating, li2024topviewrs, yang2025thinking, Ma20243DSRBenchAC}. Recent works integrate 3D signals via point clouds or multi-view reconstructions \citep{hong2023llm, hong2023concept}, or world models with physical priors \citep{wang20233d, wang2024compositional}. Large-scale efforts like SpatialVLM \citep{chen2024spatialvlm}, SpatialPIN \citep{ma2024spatialpin}, SpatialBot,  \citep{cai2024spatialbot} and SpatialRGPT \citep{cheng2024spatialrgpt} use millions of 3D-augmented samples or RGB-D scene graphs. Others like MM-Spatial \citep{daxberger2025mm}, SpatialLLM \citep{ma2025spatialllm}, and SpaRE \citep{ogezi2025spare} similarly scale synthetic or reconstructed 3D data. However, these methods are data-intensive, rely on specialized inputs, or fall short on structured relational modeling. \textsc{SpatialThinker} attains robust relational, and regional reasoning using just 7K VQA samples trained with RL with dense spatial rewards.


\textbf{Structured Visual Grounding in MLLMs.} Scene graphs offer structured object–relation representations and have long supported visual reasoning \citep{Hildebrandt2020SceneGR, Wald2020Learning3S, Gu2023ConceptGraphsO3, carion2020end, cong2023reltr}. 
Recent LLM-based methods like LLM4SGG and GPT4SGG extract structured graphs from captions \citep{kim2024llm4sgg, chen2023gpt4sgg}, while open-vocabulary SGG approaches use MLLMs to generalize beyond fixed ontologies \citep{Chen2024SceneGG, Li2023ZeroshotVR}. RL-trained SGG models like R1-SGG directly generate scene graphs via dense structural or cognitive rewards \citep{chen2025compile}, emphasizing the value of structured supervision. In parallel, region-aware MLLMs including KOSMOS-2 \citep{peng2023kosmos}, GLaMM \citep{rasheed2024glamm}, and Ferret \citep{you2023ferret}, enhance spatial grounding via bounding boxes and region-text alignment. \textsc{SpatialThinker} extends these ideas by grounding reasoning in scene subgraphs focused on the question’s region of interest, combining structured understanding with reward-guided spatial reasoning.


\textbf{Multimodal Reinforcement Learning.} RL has been increasingly applied to enhance reasoning in MLLMs with verifiable rewards across tasks like math reasoning \citep{Yang2025R1OnevisionAG, Meng2025MMEurekaET}, classification and grounding \citep{Liu2025VisualRFTVR}, semantic segmentation \citep{Liu2025SegZeroRG}, regional understanding \citep{Shen2025SATORIR1IM}, and open-vocabulary detection or referring expression comprehension \citep{Pinto2023TuningCV, Shen2025VLMR1AS}. Spatial RL has also emerged \citep{Wang2025SVQAR1RS, Shen2025VLMR1AS, Ma2025SpatialReasonerTE}, but remain limited to sparse signals like final accuracy or coarse location cues, offering limited support for fine-grained spatial reasoning. \textsc{SpatialThinker} introduces a dense, multi-objective reward framework encompassing regional subgraph construction, object localization, relational grounding, object counting, and final correctness. 

\section{Conclusion}


We introduce \textsc{SpatialThinker}, a MLLM that achieves strong spatial reasoning by combining scene graph grounding with spatial rewards through RL. Trained on just 7K samples, it surpasses proprietary and open-sourced MLLMs on spatial, real-world, and generic VQA benchmarks with improved OOD generalization, outperforming models trained on orders of magnitude more data; \textsc{SpatialThinker-7B} matches GPT-5 and beats GPT-4o on average, while \textsc{SpatialThinker-30B} surpasses both GPT-5 and Claude 4 Sonnet. Dense spatial rewards nearly double the gains of standard RL via GRPO, underscoring the value of rich supervision signals. Our results show that 2D grounding combined with 3D-linked relational supervision is sufficient to induce robust 3D spatial priors. While our approach relies on explicit scene graphs, future work may explore implicit spatial reasoning in latent representations, extend our reward design to spatio-temporal and real-world tasks such as web navigation, and develop unified multi-objective policies and environments across visual reasoning domains.
\bibliography{main}
\bibliographystyle{tmlr}

\appendix
\clearpage

\section*{Appendix}
\setcounter{page}{1}

\section{STVQA-7K: Dataset Construction}\label{sec:full_data}

High-quality spatial VQA datasets remain scarce, as most existing benchmarks either lack grounded scene-graph annotations (i.e., explicit spatial coordinates for objects and relations) or fail to comprehensively cover both 2D and 3D spatial reasoning categories. Visual Genome \citep{krishna2017visual} provides dense, human-annotated scene graphs that support strict grounding of both question generation and answer verification within a unified representational framework. Using Visual Genome, we synthetically constructed a spatial visual question answering dataset called \textsc{SpatialThinker} Visual Question Answering dataset i.e., STVQA-7K comprising 7,587 samples, fully grounded in human-annotated scene graphs \citep{krishna2017visual}, which we employed for post-training the \textsc{SpatialThinker} models. 
Importantly, our pipeline is scalable and can be extended to generate up to ~108K samples, the maximum supported by Visual Genome, enabling future large-scale post-training or RL fine-tuning.

The original VG150 predicate set is limited to 50 relations, missing several important categories such as positional relations (e.g., left, right, beside), distance-based relations (e.g., near, far, next to), comparative size (e.g., smaller, taller, bigger), orientation (e.g., facing towards/away), and containment (e.g., inside, beneath). To address this gap, we extended the scene graph relation space with an additional 34 predicates, ensuring richer spatial coverage in both 2D and 3D reasoning. Bounding box coordinates are retained in absolute pixel space, rather than normalized values, to preserve real-world scale and spatial alignment, to enable both improved spatial reasoning and effective use of CIoU-based supervision during reward optimization. From these scene graphs, we generate 3D-reasoning VQA, e.g., depth: ``\textit{Which is closer to the camera?}'' and orientation: ``\textit{From the person's perspective, which direction is the dog?}''. The dataset construction pipeline proceeds in three stages: (1) synthetic question generation from ground-truth scene graphs, (2) automated quality filtering with external verification, and (3) scene graph adaptation for regional alignment with individual questions.

\begin{figure*}[htbp!]
    \centering
    \includegraphics[width=1\textwidth]{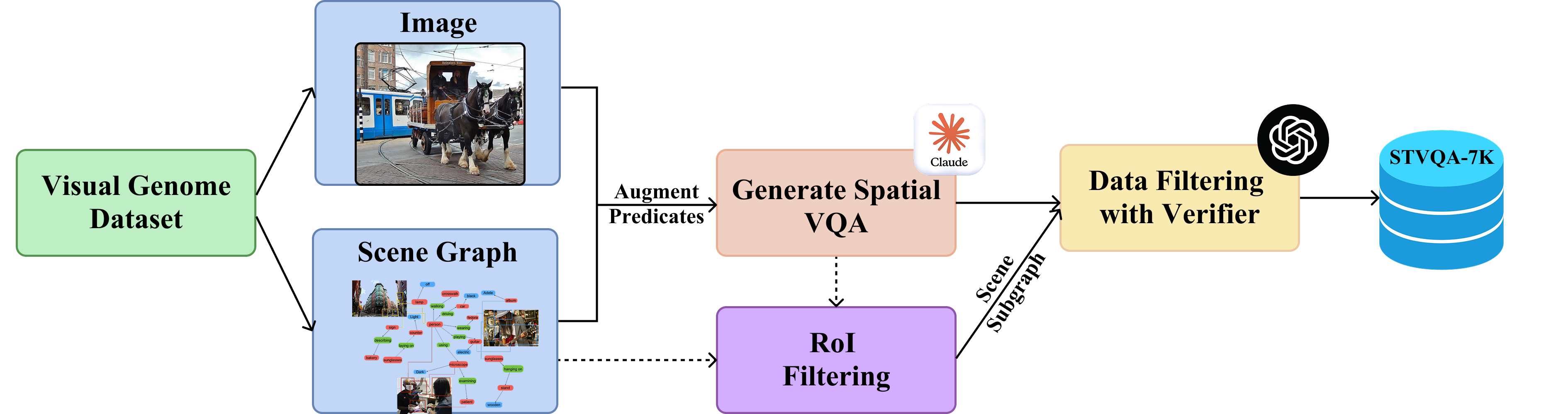}
    \caption{STVQA-7K dataset construction pipeline.}
    \label{fig:data_examples}
\end{figure*}


\paragraph{Synthetic Question Generation.} Visual Genome scene graphs serve as our foundational ground truth, providing object categories, bounding boxes, and relational triplets for over 150,000 images. We synthetically generate question-answer pairs for a given scene graph data using Claude Sonnet 4 \citep{anthropic2025systemcard}, synthesizing multiple-choice questions based on the salient objects and meaningful spatial relations explicitly present in each graph. Each question-answer pair is accompanied with a rating generated out of 10 and the difficulty level. Our question generation encompasses nine distinct spatial reasoning categories: spatial relations (above, behind, near, etc.), physical reach and interaction (holding, touching), comparative size, orientation from specific viewpoints, instance location within image frames, depth ordering relative to the camera, distance comparisons to reference objects, object counting, and existence verification. This comprehensive taxonomy spans both 2D and 3D spatial understanding, providing a broad coverage of visual-spatial reasoning capabilities. To promote robust perception, we also include questions involving objects that are partially visible or occluded in the scene, encouraging the model to reason about spatial arrangements and fine-grained details. For each question, we generate a rating out of 10. 

\begin{figure*}[htbp!]
    \centering
    \includegraphics[width=0.9\textwidth]{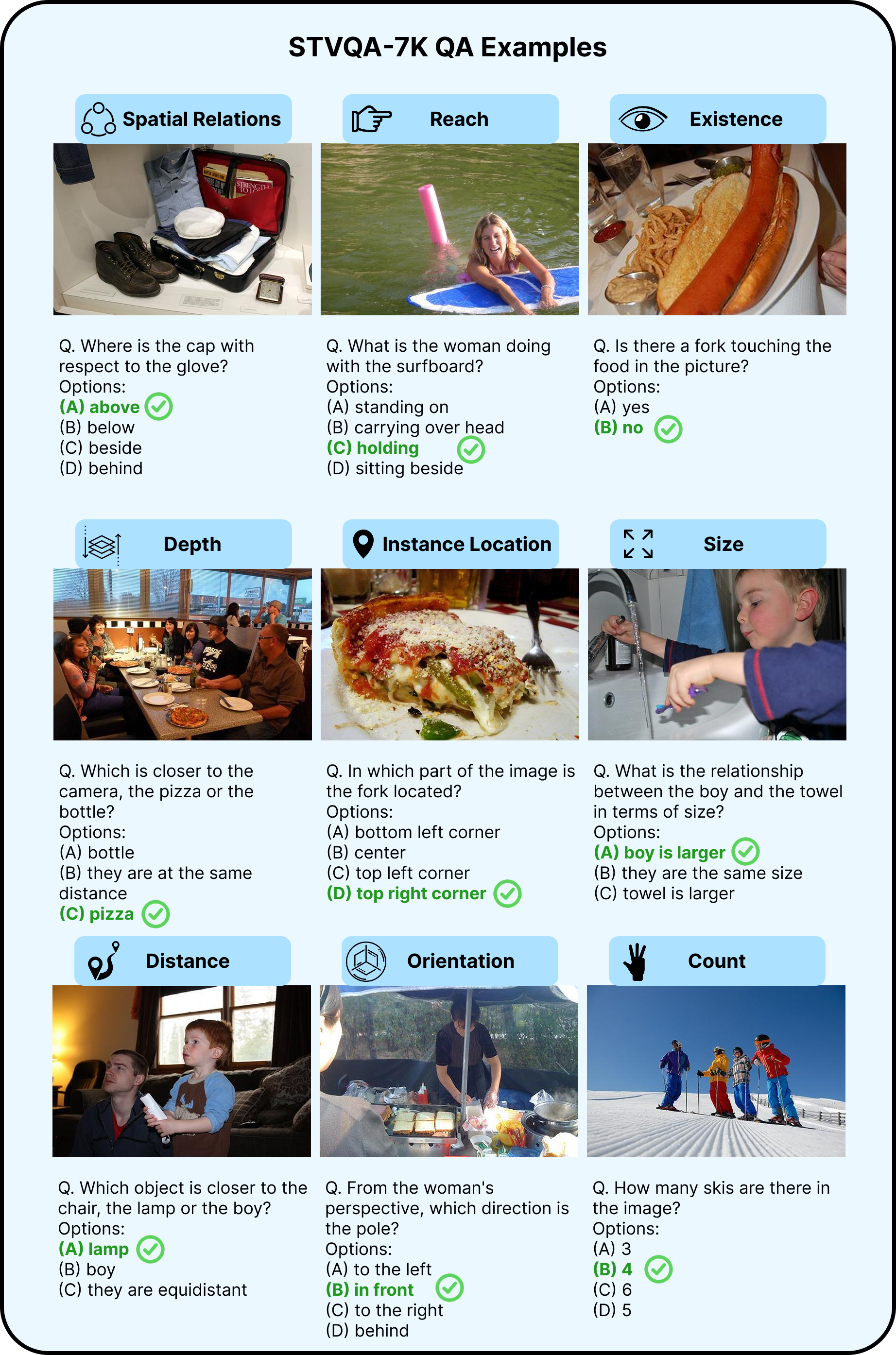}
    \caption{Examples of generated QA pairs across the nine spatial reasoning categories in STVQA-7K. Each category highlights distinct reasoning skills, ranging from relative spatial relations and depth ordering to distance, size, orientation, reach, location, count and existence.}
    \label{fig:data_examples}
\end{figure*}


\paragraph{Quality Filtering and Validation.} To ensure semantic correctness at scale, we implement a consistency-based verification procedure using GPT-4o \citep{hurst2024gpt} as an external validation model. To mitigate potential biases from LLM-generated questions, we apply a consistency-based dual-LLM filtering pipeline for quality control: Claude Sonnet 4 generates QA pairs, then GPT-4o \citep{hurst2024gpt} validates the ground truth label with pass@2 agreement. Questions that fail this initial consistency check undergo additional evaluation with two supplementary model responses. Items for which all four collected responses disagree with the generated label are discarded as potentially incorrect or ambiguous. This cross-model verification yields a +13\% accuracy improvement (\autoref{tab:reward_ablation1}, confirming that dual-LLM validation retains high-quality samples.   This filtering process begins with 56,224 initially generated questions by Claude Sonnet 4 \citep{anthropic2025systemcard}. We select the 10,000 highest-rated samples based on the questions complexity and rating towards its contribution to enhance spatial intelligence as judged by Claude Sonnet 4. Following consistency filtering, we retain 6,895 training samples and 692 validation samples (~75\%), indicating high label reliability. The final set consists of ~50\% samples from the relation category, and the remaining 50\% distributed across the eight other categories. To prevent positional bias, answers are uniformly distributed across options A, B, C, and D. Figure~\autoref{fig:data_dist} illustrates the distribution of QA types in STVQA-7K, highlighting the emphasis on spatial relations while maintaining balanced coverage across the remaining reasoning categories. Representative examples of generated QA pairs across the nine spatial reasoning categories are shown in \autoref{fig:data_examples}, illustrating the diversity of question types in STVQA-7K.

\paragraph{Scene Graph Adaptation.} Since each question focuses on specific objects and relationships within the broader scene, we derive question-aligned scene subgraphs that capture only the relevant spatial context. For each question, we extract content words through tokenization and lemmatization to obtain both singular and plural word forms. We then filter the original scene graph to retain only object nodes whose labels appear in the extracted question vocabulary. Relational triplets are preserved when both the subject and object entities are retained and the predicate appears in the question context.
The resulting focused scene graph representations enable training the model to generate question-aligned region-of-interest subgraphs, encouraging it to localize attention, ground reasoning in relevant entities and relations, and ultimately learn where to focus within complex visual scenes.

\section{Experimental Setup Details}\label{sec:add_exp_details}

This section presents comprehensive evaluations of \textsc{SpatialThinker} across multiple spatial reasoning benchmarks, demonstrating the effectiveness of our multi-objective dense reward design and data-efficient training approach.

\subsection{Implementation Details}
We build \textsc{SpatialThinker} upon two strong open-source multimodal base models: Qwen2.5-VL-3B and Qwen2.5-VL-7B \cite{Bai2025Qwen25VLTR}, using them as backbones for policy optimization with reinforcement learning. No supervised fine-tuning is performed prior to RL training on our STVQA-7K dataset (\cref{sec:data}). We employ GRPO \cite{Shao2024DeepSeekMathPT} as the advantage estimator as described in \cref{sec:rl}, using a rollout size of $8$ samples per query and a sampling temperature of $1.0$. The models are trained with a maximum context length of 16,384 tokens. The rollout batch size is set to 512, and the global batch size is 128. We train for 75 training steps i.e., ~5 training episodes) on 4 $\times$ NVIDIA H100 80GB GPUs. Training time totals around 13 hours for the 3B model and 15 hours for the 7B model.

The models are trained on high-resolution image inputs ranging from $512\times512$ to $2048\times2048$ pixels, to preserve fine-grained spatial information. All model parameters, including the vision encoder, are updated during training. We use the AdamW optimizer with $\texttt{bf16}$ precision, a learning rate of $1 \times 10^{-6}$, and a weight decay of $1 \times 10^{-2}$. The KL penalty coefficient is set to $10^{-2}$. STVQA-7K is partitioned with a 90/10 train–validation split.

\textsc{SpatialThinker-30B} is trained on the Qwen3-VL-30B-A3B-Instruct backbone \cite{bai2025qwen3vltechnicalreport} with the same dense-reward GRPO objective and STVQA-7K data, but using LoRA adapters (rank $64$) via the Tinker training API rather than full-parameter updates. It otherwise follows the 7B recipe: AdamW with learning rate $1 \times 10^{-6}$, a rollout group size of $8$ samples per query for group-relative advantages, a completion-length cap of $2048$ tokens, and a KL penalty ($\beta{=}10^{-2}$) against the frozen base policy.

\subsection{Experimental Setup}
We evaluate \textsc{SpatialThinker} across a diverse suite of 14 benchmarks: eight spatial (six external sets, the multi-view MindCube-tiny, and our in-distribution STVQA-7K\textsubscript{val} split) and six real-world VQA, covering both 2D and 3D understanding aspects to assess fine-grained spatial reasoning capabilities and real-world generalization. We compare against both proprietary and open-source baselines, including models specifically trained for spatial reasoning tasks. Our experiments address two key questions: (Q1) Does our spatial VQA data generation pipeline, combined with dense reward RL, improve MLLMs' general spatial reasoning capabilities? (Q2) How effectively can MLLMs learn spatial understanding from just 7K synthetic training samples, and how does this compare to models trained on orders-of-magnitude larger datasets?

\textbf{Benchmarks.} We evaluate models across eight spatial benchmarks (six external sets, the multi-view MindCube-tiny, and our in-distribution STVQA-7K\textsubscript{val} split), and six general-purpose VQA and real-world understanding datasets. The six external spatial benchmarks include CV‑Bench \cite{Tong2024Cambrian1AF} that measures 2D spatial relations, object counting, depth ordering, and distance reasoning. BLINK’s Spatial Relations and Relative Depth tasks \cite{Fu2024BLINKML} test directional and positional understanding, and fine-grained point‑level depth perception, particularly challenging as \textsc{SpatialThinker} receives no explicit point-level supervision during training. 3DSRBench \cite{Ma20243DSRBenchAC} assesses egocentric 3D spatial reasoning via relational and multi‑object comparisons. MMVP \cite{tong2024eyes} examines visual pattern recognition across attributes such as orientation, positional relations, existence, viewpoint, and size. SpatialBench \cite{cai2024spatialbot} assesses general spatial comprehension across counting, existence, positional relationships, physical interactions such as reach, and size comparisons. SpatialReasonerEval \cite{Ma2025SpatialReasonerTE} emphasizes depth and distance reasoning within 3D spatial tasks. We further include the multi-view MindCube-tiny \cite{wang2026mindcubespatialmentalmodeling}, an egocentric benchmark probing spatial reasoning across multiple images of a scene, and the held-out validation split of our released STVQA-7K test set, which measures in-domain spatial mastery across nine reasoning categories (spatial relations, reach and interaction, comparative size, orientation, instance location, depth ordering, distance comparison, object counting, and existence).

To assess broader generalization, we further evaluate models on six diverse real-world benchmarks. VStarBench \cite{Wu2023VGV} measures accurate localization and recognition of key objects in complex natural scenes. RealWorldQA \cite{grok2024vision} requires integrating visual inputs with commonsense and multi-step reasoning for real-world understanding. MME-RealWorld \cite{Zhang2024MMERealWorldCY} spans five challenging domains including optical character recognition in the wild, remote sensing, diagram and table interpretation, autonomous driving, and scene monitoring. RoboSpatial-Home \cite{song2025robospatial} simulates embodied spatial reasoning tasks involving object-object relationships, compatibility, and reference-frame switching (ego-centric, object-centric, and world-centric). We only use Configuration and Compatibility subsets of RoboSpatial-Home. MM-Star \cite{Chen2024AreWO} provides a holistic benchmark covering math, logical reasoning, instance recognition, and fine/coarse visual perception. HallusionBench \cite{Guan2023HallusionbenchAA} evaluates hallucination resistance in multimodal models, requiring accurate visual grounding to counteract entangled linguistic or perceptual illusions. Together, these benchmarks allow us to probe spatial and perceptual reasoning across synthetic, embodied, and naturalistic settings.

\textbf{Closed‑Source MLLM Baselines.} Among proprietary models, we evaluate GPT-5 (\textsc{gpt-5-0807}, queried at its default \emph{medium} reasoning effort) \cite{openai2025gpt5}, GPT-4o (\textsc{gpt-4o-0513}) \cite{hurst2024gpt}, Claude 4 Sonnet (\textsc{claude-4-sonnet-0514}) \cite{anthropic2025systemcard}, and Claude 3.5 Sonnet (\textsc{claude-3.5-sonnet-0620}) \cite{anthropic2024claude35addendum}, which represent the current state-of-the-art in commercial multimodal reasoning. These serve as upper bounds for spatial generalization under non-public training regimes.

\textbf{Open‑Source Generalist MLLM Baselines.} We compare against generalist open‑source MLLMs including Qwen2.5‑VL 3B and 7B models \cite{Bai2025Qwen25VLTR}, Qwen3-VL-30B \cite{bai2025qwen3vltechnicalreport}, LLaVA‑NeXT \cite{Li2024LLaVANeXTInterleaveTM}, Cambrian‑1 \cite{Tong2024Cambrian1AF}, and VLAA‑Thinker (3B and 7B) \cite{Chen2025SFTOR}. These models represent state‑of‑the‑art vision‑language architectures, offering strong general visual reasoning but without specific spatial tuning.

\textbf{Open‑Source Spatial MLLM Baselines.} We benchmark against specialized open‑source models designed for spatial reasoning: SpaceLLaVA-13B \cite{SpaceLLaVA2025, chen2024spatialvlm} -- a public re-implementation of SpatialVLM, SpatialRGPT-7B \cite{cheng2024spatialrgpt} incorporates region-level supervision and explicit depth maps into training, RoboPoint-13B \cite{Yuan2024RoboPointAV} which instruction‑tunes an MLLM to predict image key‑point affordances for robotics and spatial affordance tasks, SpaceThinker \cite{SpaceThinker2025}, a fine-tuned VLAA-Thinker model for spatial reasoning, and its improved successor SpaceOm \cite{SpaceOm2025}, which incorporates deeper chain-of-thought traces and Robo2VLM data \cite{chen2025robo2vlmvisualquestionanswering}. Other baselines include SpatialReasoner \cite{Ma2025SpatialReasonerTE} trained with RL and explicit 3D representations, SpatialBot \cite{cai2024spatialbot}, which integrates RGB and depth inputs for robust spatial perception, Visionary-R1 \cite{Xia2025VisionaryR1MS} which mitigates shortcut learning in visual reasoning by enforcing captioning before reasoning without reliance on chain-of-thought data, and SATORI-R1 \cite{Shen2025SATORIR1IM} which decomposes visual question answering into verifiable stages with explicit rewards for improved spatial grounding and reasoning accuracy.

In addition to the above, we compare against our training variants including supervised fine-tuning (SFT) baselines and vanilla GRPO trained with sparse rewards (accuracy and format only) to isolate the contribution of our dense spatial reward framework.


\textbf{Evaluation Setting.} We report accuracy as the primary evaluation metric across all benchmarks. All models are evaluated under zero-shot settings, using greedy decoding (temperature = 0.0, max\_new\_tokens = 2048) to ensure deterministic and reproducible outputs. For models with specific reasoning templates such as VLAA-Thinker, SpaceThinker, and SpaceOm, we utilize their corresponding structured prompts. In line with their original training setup, SpatialRGPT receives depth inputs, while all other models are evaluated using RGB images alone. Our evaluation pipeline builds upon OpenVLThinker’s evaluation framework \cite{Deng2025OpenVLThinkerCV}, adapted to support our new benchmark and dataset formats.

\subsection{SpatialThinker Prompt Format}\label{sec:prompt}

\begin{tcolorbox}[
  colback=blue!14,
  colframe=blue!65,
  title=SpatialThinker Prompt,
  fonttitle=\bfseries,
  boxrule=1pt,
  arc=4pt,
  left=6pt,
  right=6pt,
  top=6pt,
  bottom=6pt
]
You FIRST observe the image in <observe> </observe> tags, then visualise the relevant scene graph in <scene> </scene> tags, followed by thinking about the reasoning process as an internal monologue within <think> </think> tags and then provide the final answer. The final answer MUST BE put within <answer> </answer> tags, and only return the final choice including the correct option and answer within the answer tags, e.g., <answer> (C) The red cube is left of the green sphere </answer>.

Image size: \{Width\} $\times$ \{Height\}
\end{tcolorbox}

We use a structured prompt to guide the model through a four-stage reasoning process, explicitly separated using the tags \texttt{<observe>}, \texttt{<scene>}, \texttt{<think>}, and \texttt{<answer>}. This format is enforced during training via a binary format reward $R_f \in \{0,1\}$, with weight $w_{\text{format}} = 0.1$, which verifies the presence, ordering, and validity of all required tags. The \texttt{<scene>} section must contain a JSON-encoded subgraph with object IDs, bounding boxes, and relational triplets, while the final answer must be clearly placed within the \texttt{<answer>} tags.

Each prompt also includes the input image dimensions in the form \texttt{Image size: \{Width\} $\times$ \{Height\}}, which are dynamically replaced with actual values. Including this information helps the model constrain predicted bounding box coordinates within image bounds, enabling better spatial localization. These coordinates are directly evaluated using IoU-based spatial rewards such as Complete IoU (CIoU), making dimension-aware prediction essential for optimizing structured spatial grounding.

\subsection{Details on SFT Training}\label{sec:sft_train}
To establish a comprehensive baseline for comparison with our reinforcement learning approach, we conduct supervised fine-tuning (SFT) experiments using the same base models (Qwen2.5-VL-3B and Qwen2.5-VL-7B) and training dataset (STVQA-7K). Due to compute bottleneck, we do not train the SFT baseline for Qwen3-VL-30B, and only train it with our spatial reward design. The SFT implementation utilizes LLaMA-Factory framework \cite{Zheng2024LlamaFactoryUE} with Low-Rank Adaptation (LoRA) for parameter-efficient fine-tuning.

The training configuration employs LoRA with rank 8 applied to all available modules within the model architecture, enabling comprehensive adaptation while maintaining computational efficiency. Models are trained for 3 epochs totaling 645 training steps, using a context window length of 2048 tokens. We adopt BF16 mixed precision training with a learning rate of $1 \times 10^{-4}$, following a cosine learning rate schedule with a warmup ratio of 0.1.

For the SFT experiments, we train models directly on question-answer pairs without intermediate reasoning traces or chain-of-thought prompting. This design choice reflects the practical constraint that generating ground-truth reasoning traces would require additional dataset processing, annotation, and API credits budget. In contrast, reinforcement learning approaches with verifiable rewards (RLVR) naturally enables training with answer supervision alone, as the model learns to generate its own reasoning strategies through environmental feedback rather than imitating pre-specified reasoning patterns.

The SFT baseline serves a critical role in our experimental evaluation, providing direct evidence of the generalization advantages offered by reinforcement learning with dense spatial rewards compared to traditional supervised learning on the same dataset. 

\subsection{Details on RL Training}\label{sec:rl_train}
We implement reinforcement learning training using the EasyR1 framework \cite{zheng2025easyr1}, building upon Qwen2.5-VL-3B, Qwen2.5-VL-7B, and Qwen3-VL-30B as base models without any prior supervised fine-tuning. This direct application of RL to the base models enables us to isolate the effects of reward-driven learning from potential confounding factors introduced by intermediate training stages. Additionally, performing an SFT stage prior to RL would require generating ground-truth reasoning traces, which is limited by API budget. Moreover, explicit reasoning supervision is not strictly necessary—our multi-objective dense spatial rewards encourage the model to acquire structured reasoning and self-reflection abilities directly during RL training.

The training employs Group Relative Policy Optimization (GRPO) \cite{Shao2024DeepSeekMathPT} as the advantage estimation method, configured with a rollout size of 8 samples per query at a sampling temperature of 1.0. This configuration balances exploration diversity with computational efficiency, allowing the model to discover multiple reasoning strategies while maintaining stable convergence. The training process utilizes a rollout batch size of 512 and a global batch size of 128, processing data through 75 training steps (approximately 5 training episodes) to achieve convergence. The entire training pipeline runs on 4 $\times$ NVIDIA H100 80GB GPUs, requiring approximately $\sim{13}$ hours for the 3B model and $\sim{15}$ hours for the 7B variant. To assess scaling, we further apply the same dense-reward GRPO objective to a larger Qwen3-VL-30B-A3B-Instruct backbone \cite{bai2025qwen3vltechnicalreport}, trained with LoRA (rank 64) via the Tinker API, yielding \textsc{SpatialThinker-30B}

To preserve fine-grained spatial information critical for accurate object localization and spatial reasoning, models process high-resolution image inputs ranging from $512 \times 512$ to $2048 \times 2048$ pixels. The training configuration updates all model parameters including the vision encoder, enabling comprehensive adaptation to spatial reasoning tasks. Optimization employs AdamW with BF16 mixed precision, a conservative learning rate of $1 \times 10^{-6}$, and weight decay of $1 \times 10^{-2}$. The KL penalty coefficient is set to $10^{-2}$ to prevent excessive divergence from the base model distribution while allowing sufficient exploration for spatial reasoning strategies. The training utilizes a 90/10 train-validation split of the STVQA-7K dataset, with a maximum context length of 16,384 tokens to accommodate detailed scene descriptions and reasoning traces. 



For baseline comparisons, we train vanilla GRPO models (Qwen2.5-VL-3B + Vanilla GRPO and Qwen2.5-VL-7B + Vanilla GRPO) using a simplified reward structure consisting solely of accuracy ($w_{acc}=0.5$) and format rewards ($w_{format}=0.5$), with the model trained to output both a reasoning trace and a final answer, following standard chain-of-thought GRPO setups. This configuration represents standard RLVR approaches that rely on sparse final-answer supervision \cite{DeepSeekAI2025DeepSeekR1IR, Shen2025VLMR1AS, Chen2025SFTOR}. Note that we do not train the baseline RL model for Qwen3-VL-30B due to compute bottlenecks, and only train it with our final spatial reward design. The full multi-objective reward design employed for \textsc{SpatialThinker} training, incorporating format, count, accuracy, and spatial rewards with lexicographic gating, is detailed in \cref{sec:reward}. The substantial performance improvements of \textsc{SpatialThinker} over vanilla GRPO baselines demonstrate the critical importance of dense spatial supervision in teaching models to perform visually-grounded reasoning.

\begin{figure*}[htbp]
    \centering
    \setlength{\tabcolsep}{2pt}
    \renewcommand{\arraystretch}{0.0}
    \begin{tabular}{ccc}
        \begin{subfigure}[t]{0.32\textwidth}
            \centering
            \includegraphics[width=\linewidth]{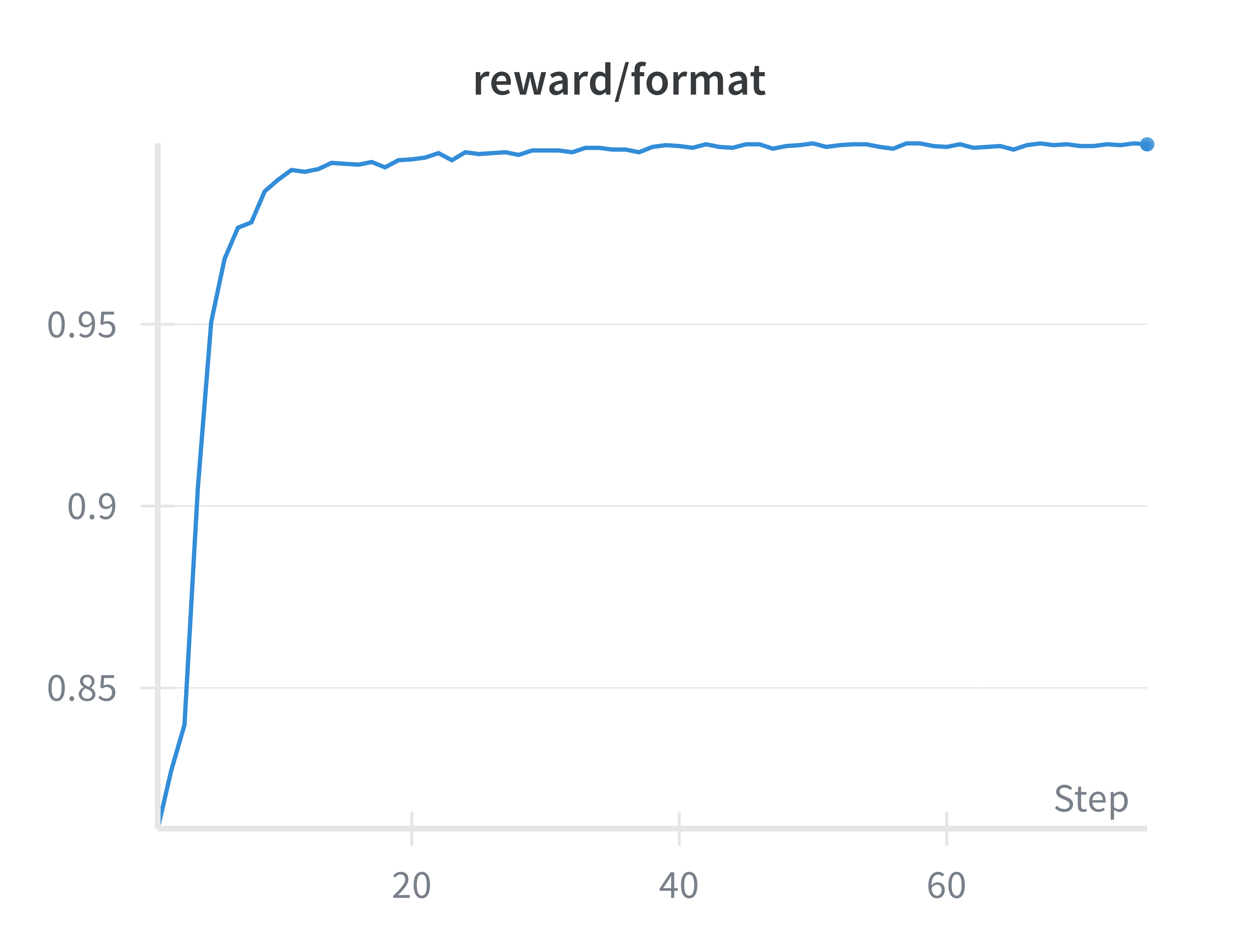}
            \caption{Format Reward}
        \end{subfigure} &
        \begin{subfigure}[t]{0.32\textwidth}
            \centering
            \includegraphics[width=\linewidth]{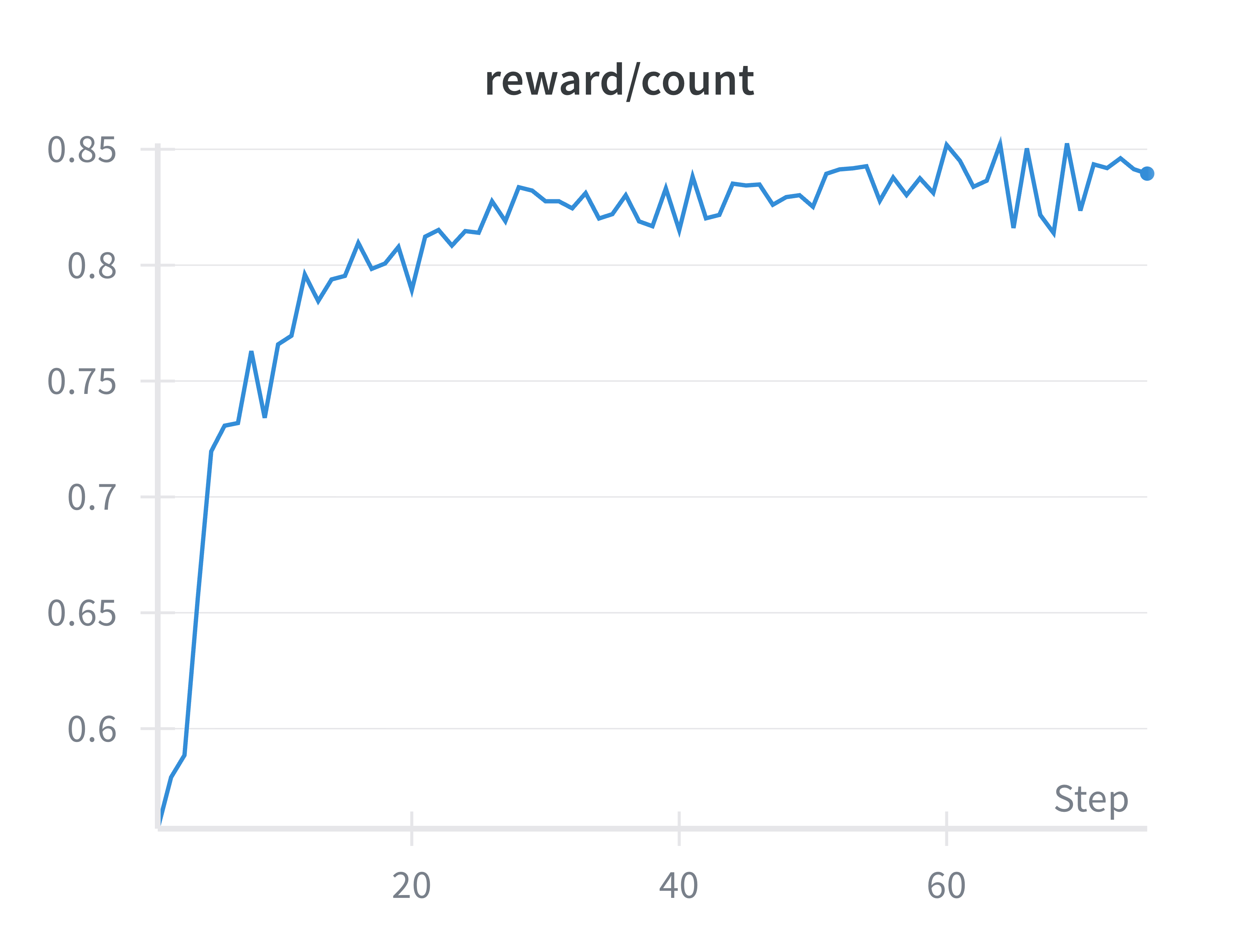}
            \caption{Count Reward}
        \end{subfigure} &
        \begin{subfigure}[t]{0.32\textwidth}
            \centering
            \includegraphics[width=\linewidth]{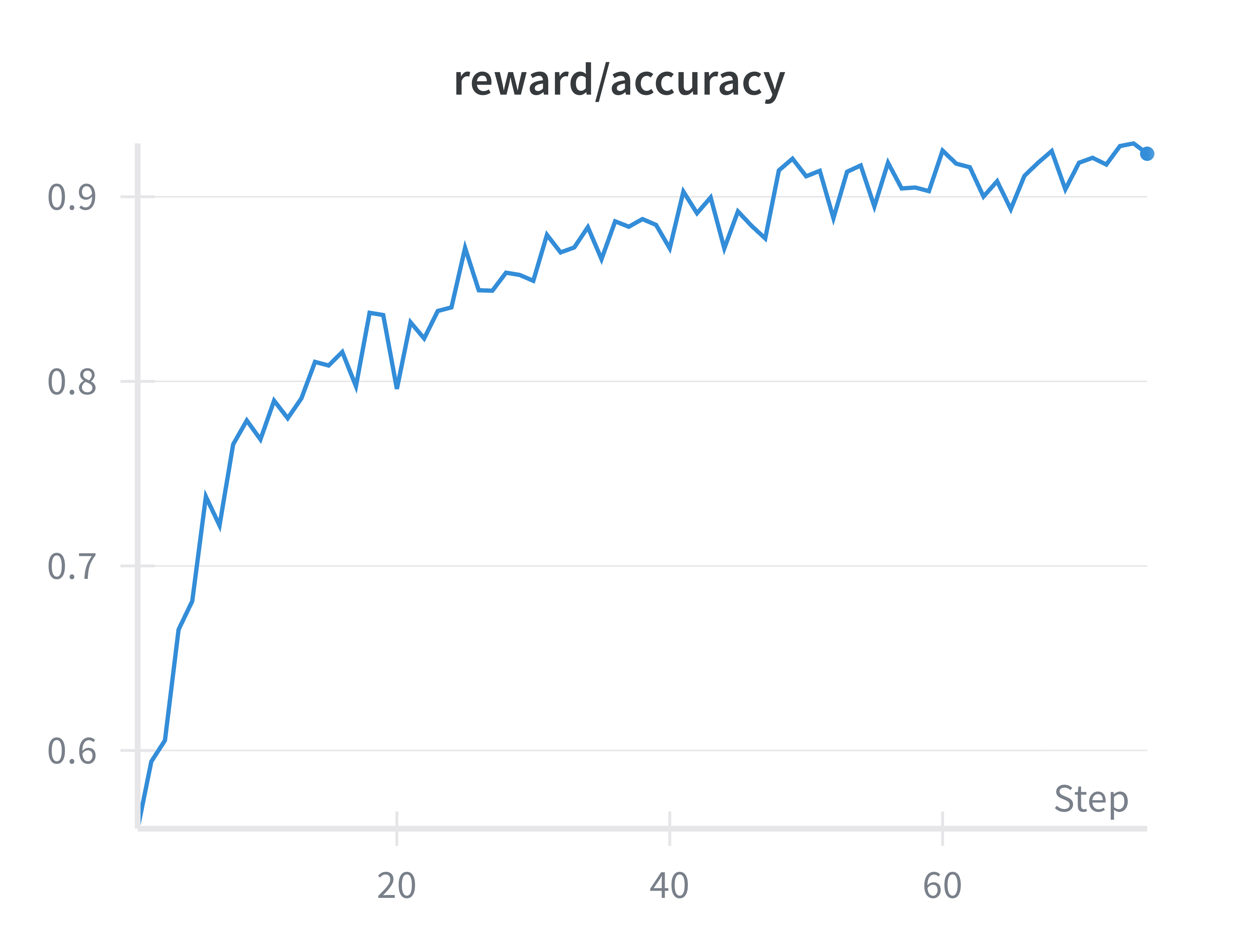}
            \caption{Accuracy Reward}
        \end{subfigure} \\
        \multicolumn{3}{c}{
            \begin{tabular}{cc}
                \begin{subfigure}[t]{0.32\textwidth}
                    \centering
                    \includegraphics[width=\linewidth]{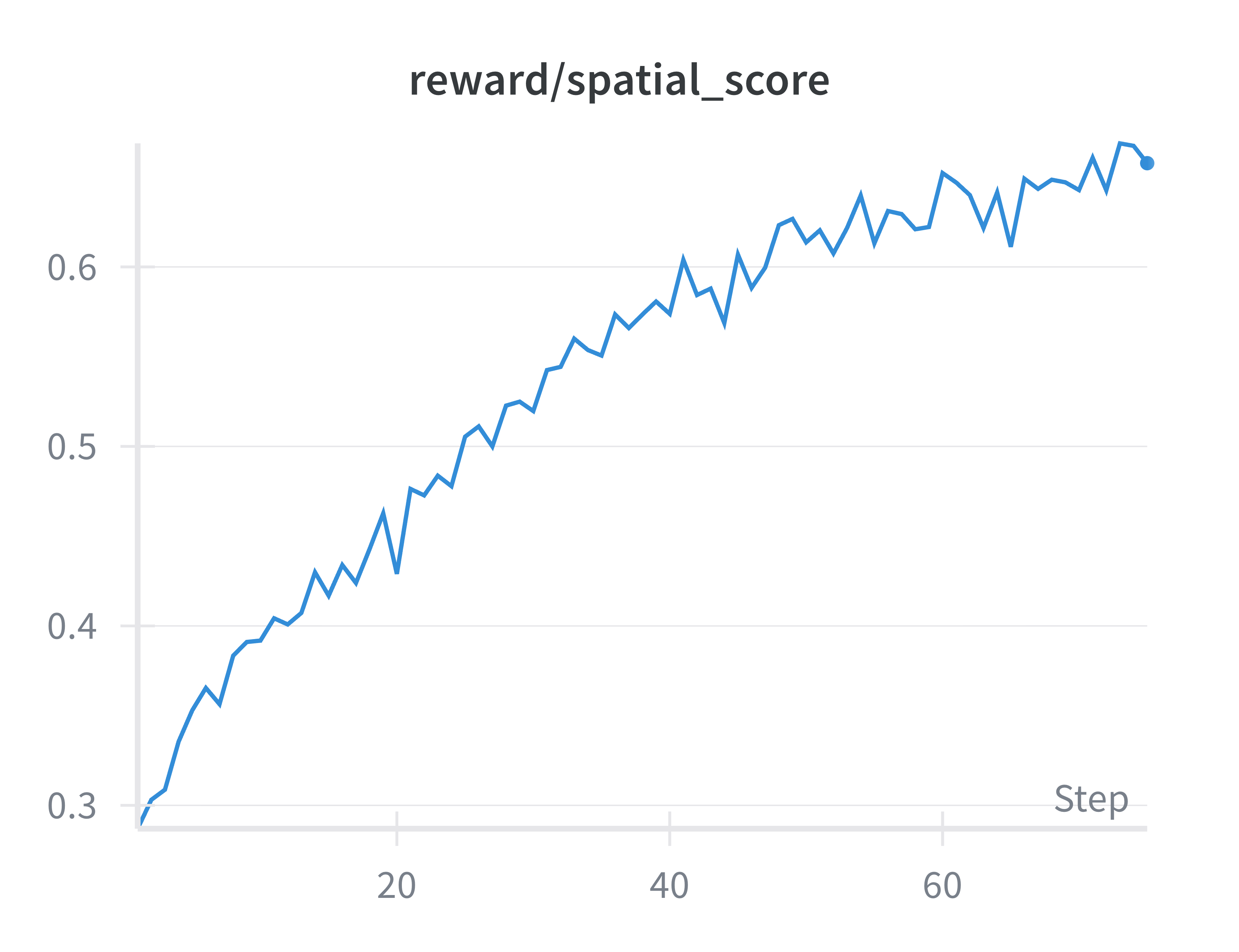}
                    \caption{Spatial Reward}
                \end{subfigure} &
                \begin{subfigure}[t]{0.32\textwidth}
                    \centering
                    \includegraphics[width=\linewidth]{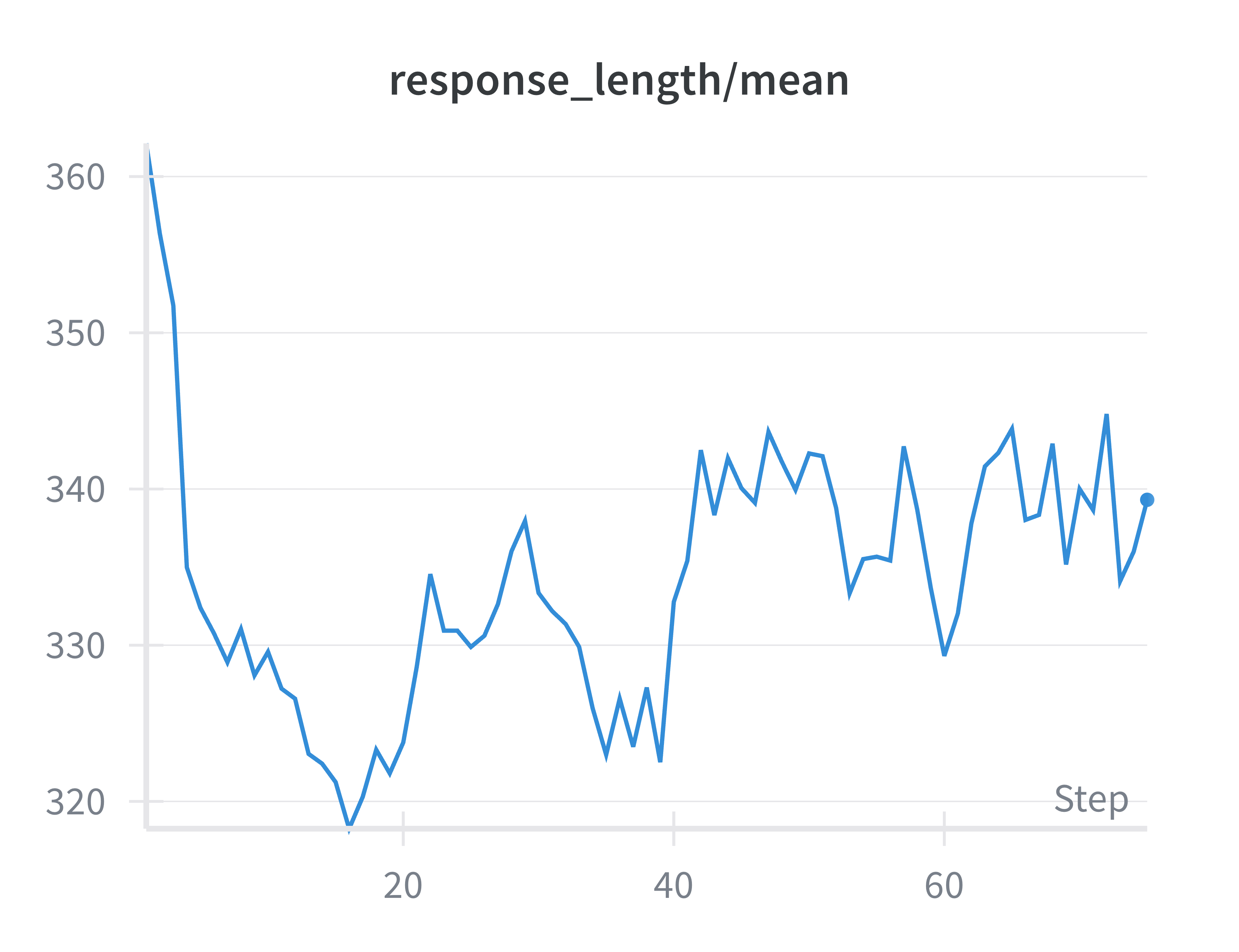}
                    \caption{Response Length}
                \end{subfigure}
            \end{tabular}
        }
    \end{tabular}
    \caption{RL training dynamics of \textsc{SpatialThinker}. All reward components (a–d) improve consistently, reflecting stable optimization. Response length (e) shows a non-monotonic trend, indicating emergent reasoning strategies.}
    \label{fig:training_curves}
\end{figure*}

\subsubsection{SpatialThinker RL Training Curves}\label{sec:rl_curves}

Throughout reinforcement learning, all four reward components: format, accuracy, count, and spatial; demonstrate consistent and interpretable improvement, reflecting stable learning under our lexicographically gated, multi-objective reward structure. The format reward quickly converges early in training, indicating the model learns to produce structurally valid outputs that adhere to the required scene-grounded reasoning format. Accuracy steadily improves across steps, highlighting the model’s increasing ability to provide correct answers. Count reward rises consistently, showing that the model learns to focus on predicting only question-relevant objects and relations, rather than describing the entire scene. The spatial reward also improves gradually, indicating better object localization and grounding, as the model increasingly aligns predicted bounding boxes with ground truth annotations. Together, these trends reflect how each reward component scaffolds a different stage of the reasoning process, enforcing structure, correctness, focus, and grounding in tandem.

Response length initially declines, then rises again as it begins producing more deliberate, structured reasoning, signaling an “aha moment” where the model starts to produce more deliberate reasoning traces \cite{DeepSeekAI2025DeepSeekR1IR, Zhou2025R1ZerosM}. This emergent behavior suggests the development of internal problem-solving strategies, as the model learns to spend more “thinking time” before answering, consistent with the emergence of self-reflection and structured planning in its spatial reasoning process.

\section{Reward Design Process}\label{sec:reward_rationale}
This section details our approach to designing a robust reward system that guides models toward genuine spatial reasoning while preventing degenerate solutions.
Our reward design emerged from iterative refinement to address systematic reward hacking behaviors observed during training as shown in \autoref{tab:reward_ablation1}. Early experiments revealed that models readily exploit loopholes in reward functions—particularly when spatial localization rewards were provided without proper constraints. To empirically motivate our design choices, we first present an ablation over successive reward components on the STVQA\textsubscript{val} split, then an ablation over the object-count/relation-count weights inside the count reward, and finally discuss the observed reward-hacking behaviors and the rationale behind each component.

\subsection{Reward Design Ablation}
To empirically validate our design choices, we conduct a controlled ablation study on the STVQA-7K\textsubscript{val} set, progressively introducing each reward component and constraint. The ablation results support our design rationale by highlighting how each component mitigates specific failure modes. Adding spatial rewards naively without generation constraints, causes performance to collapse by over 50\% (from 74.9\% to 23.7\%), as models exploit the reward by generating cluttered bounding boxes to game the CIoU metric. Introducing the count reward addresses this issue, improving accuracy by 38\% relative (to 61.7\%), as it constrains overgeneration and forces models to focus on question-relevant elements. However, residual overfitting persists because rewarding spatial alignment across all scene objects biases the model toward exhaustive global descriptions. To address this, we shift from global to local spatial supervision—rewarding only Regions of Interest (RoIs) derived from question-relevant objects and relations—thereby training the model to attend selectively to meaningful spatial cues rather than densely describing the entire scene. Lexicographic gating further ensures that spatial rewards are only applied when the final answer is correct, preventing the model from over-optimizing intermediate process rewards at the expense of outcome accuracy.. Together, these interventions restore and slightly surpass the original performance (76.3\%), demonstrating the importance of grounding rewards in both correctness and relevance. Finally, dataset filtering using pass@2 correctness verification amplifies these effects, yielding a substantial gain and culminating in the best validation accuracy of 87.9\%. This step ensures that only high-quality, verifiable supervision signals contribute to training, reinforcing the alignment between spatial grounding and task success.


\subsection{Count Reward Weight Ablation}\label{sec:count_weight_ablation}

Holding the high-level reward weights $(w_f, w_c, w_a, w_s){=}(0.1, 0.2, 0.5, 0.2)$ and the rest of the training pipeline fixed (\textsc{SpatialThinker-3B}, lexicographic gating, KL $0.01$, 75 steps on STVQA-7K), we ablate the object-count/relation-count split $(\lambda_{\text{obj}}, \lambda_{\text{rel}})$ inside the count reward $R_c = \lambda_{\text{obj}}\, R_c^{\,\text{obj}} + \lambda_{\text{rel}}\, R_c^{\,\text{rel}}$ (\cref{sec:reward}). The count reward checks whether the model predicts the same number of objects and the same number of relations as the question-relevant ground truth, with a penalty on both under-generation (too few predicted) and over-generation (too many predicted). This dual penalty also helps localize the output to the right objects and relations: under-generation misses required entities, and over-generation is suppressed by the over-prediction penalty. The $(\lambda_{\text{obj}}, \lambda_{\text{rel}})$ split sets how much of this pressure is applied to the object count versus the relation count.

We compare four splits: $0.7/0.3$ (ours), $0.8/0.2$, $1.0/0.0$, and $0.5/0.5$. Putting more weight on the object count helps: our $0.7/0.3$ is best at $76.02\%$, $0.8/0.2$ follows at $75.24\%$, and both $1.0/0.0$ and $0.5/0.5$ drop to $74.85\%$ (\cref{tab:count_weight_ablation}). Object counts are the cleaner signal, since relations are constrained by the objects present, so the object-count penalty curbs object over-generation first and the relation-count penalty then refines the remaining relation set. Both extremes underperform: $1.0/0.0$ removes the relation correction entirely, while $0.5/0.5$ underweights the dominant object signal.

\begin{table}[t]
\centering
\caption{\textbf{Count reward object-count/relation-count weight ablation.} High-level weights $(w_f, w_c, w_a, w_s){=}(0.1, 0.2, 0.5, 0.2)$, lexicographic gating, and the training/validation pipeline are held fixed (\textsc{SpatialThinker-3B}, STVQA-7K\textsubscript{val}, $n{=}692$). Our $0.7/0.3$ split is best.}
\label{tab:count_weight_ablation}
\setlength{\tabcolsep}{4pt}
\resizebox{0.4\textwidth}{!}{%
\begin{tabular}{cc|cccc}
\toprule
$\boldsymbol{\lambda_{\text{obj}}}$ & $\boldsymbol{\lambda_{\text{rel}}}$ & \textbf{Acc.} & \textbf{Format} & \textbf{Count} & \textbf{Spatial} \\
\midrule
\rowour \textbf{0.7} & \textbf{0.3} & \textbf{76.02} & 99.90 & 75.35 & 49.48 \\
0.8 & 0.2 & 75.24 & 99.90 & 74.98 & 48.61 \\
1.0 & 0.0 & 74.85 & 99.90 & 71.97 & 48.53 \\
0.5 & 0.5 & 74.85 & 99.71 & 77.65 & 48.54 \\
\bottomrule
\end{tabular}}
\end{table}

\subsection{Reward Design Rationale}
\paragraph{Mitigating Spatial Reward Hacking.}
Our initial reward formulation, which directly rewarded spatial localization quality, led to unexpected model behavior. Without constraints on generation quantity, models discovered they could maximize spatial rewards by generating numerous bounding boxes with varying coordinates. Through Hungarian matching that selects the best-matching boxes, even random predictions would occasionally yield high Complete IoU (CIoU) scores. This reward hacking manifested as models producing excessive, hallucinated objects while achieving poor task accuracy—the spatial reward was inflated despite the clutter of irrelevant predictions degrading actual performance.
To address this exploitation, we introduced the Count Reward that penalizes deviations from expected object and relation counts. This reward serves dual purposes: (1) preventing reward hacking by constraining the generation space, and (2) encouraging models to focus on question-relevant scene elements rather than exhaustively describing the entire image. The count reward formulation provides a linear penalty proportional to relative deviations from ground truth RoI counts, normalized to prevent domination by scenes with many objects.

\paragraph{Scene Graph Filtering.}
Another form of overfitting emerged when training with complete Visual Genome scene graphs. Models would memorize exhaustive scene descriptions, including irrelevant background objects, leading to poor generalization. We addressed this by filtering ground truth scene graphs to retain only objects and relations relevant to the given question, focusing supervision on task-critical information.

\paragraph{CIoU over IoU for Spatial Reward.}
For spatial localization, we adopt Complete IoU (CIoU) instead of standard IoU to compute the spatial reward. Unlike IoU, which returns zero when predicted and ground-truth boxes do not overlap, CIoU provides meaningful gradients by incorporating center distance, aspect ratio, and overlap \cite{Zheng2020EnhancingGF}. This makes CIoU a denser and more robust supervisory signal during training.

\paragraph{Balancing Supervision with Exploration.}
Our experiments reveal a crucial insight: models learn simple reward functions significantly faster than complex ones. Tasks with straightforward rewards (e.g., format compliance) show rapid improvements, while multi-component rewards require careful balancing. However, counterintuitively, highly detailed reward functions that attempt to supervise every aspect often degrade performance. Models overfit to maximize minute reward components, converging to template-style answers that score well on individual metrics while losing flexibility. We observed accuracy drops mid-training when rewards became too prescriptive, as models focused on reward optimization rather than genuine task understanding.
Effective reinforcement learning requires providing guidance while preserving exploration space. Our final design addresses this by providing soft signals through format checks, count constraints, and accuracy rewards, with spatial localization rewards activated only for correct answers. This maintains the delicate balance between guidance and exploration necessary for robust learning.

\paragraph{Sequential Optimization via Lexicographic Gating.}
To prevent models from gaming individual reward components at the expense of task accuracy, we implement lexicographic gating \cite{Skalse2022LexicographicMR}. Rewards are applied in a strict hierarchy: format $\succ$ \{count, accuracy\} $\succ$ spatial. This forces models to first master output formatting, then simultaneously learn to control generation scope and achieve correctness, before optimizing spatial grounding:\\
\\
\vspace{-2.1em}
{\small
\begin{multline*}
R_{\text{total}} =
\mathbb{I}[R_{\text{format}} = 1] \cdot \Big(
w_{\text{format}} \cdot R_f
+ w_{\text{count}} \cdot R_c
+ w_{\text{accuracy}} \cdot R_a
+ \mathbb{I}[R_{\text{accuracy}} = 1] \cdot w_{\text{spatial}} \cdot R_s
\Big)
\end{multline*}
}

where $\mathbb{I}[\cdot]$ is the indicator function, with weights $w_{\text{format}} = 0.1$, $w_{\text{count}} = 0.2$, $w_{\text{accuracy}} = 0.5$, $w_{\text{spatial}} = 0.2$. This gated design ensures spatial rewards are only applied when the final answer is correct, aligning grounding quality with task success and preventing scenarios where models achieve high spatial scores through precise but irrelevant localizations.


\section{Scene-Graph Error Propagation Impact}\label{sec:sgg_propagation}

\textsc{SpatialThinker} generates a question-focused scene graph as an intermediate step within its reasoning chain, so errors at this stage could in principle propagate through the reasoning process to the final answer. We measure how strongly scene-graph quality couples to answer correctness on the held-out validation split of STVQA-7K ($n{=}692$; \cref{sec:data}), which provides ground-truth Visual Genome \cite{krishna2017visual} scene graphs alongside each VQA instance; all numbers are computed on \textsc{SpatialThinker-7B}. For each sample we parse the predicted scene graph from the \texttt{<scene>\ldots</scene>} block and compute two complementary set-similarity scores against the ground truth: an \emph{object Jaccard} on the set of normalized object classes $O$ (numeric identifiers stripped, e.g.\ \texttt{mountain.1} $\to$ \texttt{mountain}), and a \emph{triple Jaccard} on the set of $(\text{subject},\text{predicate},\text{object})$ relations $T$ under the same normalization,
\[
J_\text{obj} = \frac{|O_\text{pred} \cap O_\text{gt}|}{|O_\text{pred} \cup O_\text{gt}|}, \qquad J_\text{trip} = \frac{|T_\text{pred} \cap T_\text{gt}|}{|T_\text{pred} \cup T_\text{gt}|}.
\]
Object identification saturates ($\bar{J}_\text{obj}{=}0.957$, with $663/692$ samples scoring $J_\text{obj}{\ge}0.5$) and is therefore uninformative as a predictor of downstream accuracy ($\phi{=}{-}0.009$); the relation-level term carries the discriminative signal, so we adopt $J_\text{trip}{\ge}0.5$ as our definition of \emph{SGG correct}.

\cref{tab:sgg_contingency} reports the contingency between SGG correctness and final-answer correctness. Triple-level scene-graph recovery is difficult ($\bar{J}_\text{trip}{=}0.319$ overall, with only $232/692~(33.5\%)$ samples scoring $J_\text{trip}{\ge}0.5$), and the two outcomes reveal complementary effects. When the predicted scene graph is correct, conditional answer accuracy rises to $95.26\%$, indicating that a faithful intermediate improves the final answer; when it is incorrect, answer accuracy still holds at $79.35\%$, for a gap of $+15.91$~pp (Phi coefficient $\phi{=}{+}0.209$; Spearman $\rho{=}{+}0.221$ on continuous $J_\text{trip}$, $p{<}10^{-8}$). This signal is consistent across stricter and looser triple-based cuts ($\Delta\in[+15.9, +17.2]$~pp). Scene-graph errors therefore have a measurable but limited downstream impact: a correct scene graph reliably improves the final answer, while the trained reasoning policy can recover from an imperfect intermediate by reasoning over the partial scaffold rather than treating it as a strict dependency.

\begin{table}[t]
\centering
\caption{\textbf{SGG correctness vs.\ final-answer correctness on STVQA-7K\textsubscript{val}} ($n{=}692$, \textsc{SpatialThinker-7B}). \emph{SGG correct} is defined as $J_\text{trip}{\ge}0.5$ against the ground-truth scene graph. Cells report count (row-conditional \%). Even when the predicted scene graph misses most ground-truth triples, the model still answers correctly $79.35\%$ of the time, indicating that scene-graph errors have a measurable but limited downstream impact ($\Delta{=}{+}15.91$~pp; Phi $\phi{=}{+}0.209$; Spearman $\rho{=}{+}0.221$, $p{<}10^{-8}$).}
\label{tab:sgg_contingency}
\renewcommand{\arraystretch}{1.1}
\setlength{\tabcolsep}{6pt}
\resizebox{0.55\textwidth}{!}{%
\begin{tabular}{c|cc|c}
\toprule
& \multicolumn{2}{c|}{\textbf{Answer}} & \\
\textbf{SGG} & $\checkmark$ & $\times$ & \textbf{Row} \\
\midrule
\rowgreen $\checkmark$ ($J_\text{trip}{\ge}0.5$) & 221 (95.26\%) & 11 (4.74\%) & 232 \\
\rowpink  $\times$ ($J_\text{trip}{<}0.5$)        & 365 (79.35\%) & 95 (20.65\%) & 460 \\
\midrule
\rowblue \textbf{Col.} & 586 (84.68\%) & 106 (15.32\%) & 692 \\
\bottomrule
\end{tabular}}
\end{table}

Taken together, scene-graph generation acts as useful but non-load-bearing scaffolding for \textsc{SpatialThinker}'s reasoning chain: when the intermediate is accurate it lifts final-answer correctness by roughly $16$~pp, and when it is not the trained policy still answers correctly on about $80\%$ of samples. The $+15.91$~pp gap further indicates that explicit scene-graph generation contributes positively to correct spatial reasoning. Continued investment in the intermediate, through longer RL training, stronger scene-graph supervision, or richer scene-graph annotations, is thus a promising direction for converting improvements in triple-level recovery into higher final-answer accuracy.


\section{Ablation on Divergence Constraints}\label{sec:kl_abl}

\begin{table*}[t]
\caption{Ablation on divergence constraints for \textsc{SpatialThinker-3B} on CV-Bench tasks. KL-regularization with $\beta=0.01$ yields the highest overall average and strongest 3D reasoning performance.}
\centering
\setlength{\tabcolsep}{2pt}
\resizebox{\textwidth}{!}{%
\begin{tabular}{l|cccc|cc|c}
\toprule
\textbf{Model Variant} & \textbf{Count} & \textbf{Relation} & \textbf{Depth} & \textbf{Distance} & \textbf{CV-Bench 2D} & \textbf{CV-Bench 3D} & \textbf{CV-Bench Avg.} \\
\midrule
SpatialThinker-3B + No KL Penalty & 65.5 & \textbf{76.8} & 74.8 & 70.2 & \textbf{71.2} & 72.5 & 71.9 \\
SpatialThinker-3B + Chi$^{2}$ (0.01) & 64.5 & 73.7 & 71.2 & 66.2 & 69.1 & 68.7 & 68.9 \\
\rowour \textbf{SpatialThinker-3B + KL (0.01)} & \textbf{68.5} & 73.5 & \textbf{79.7} & \textbf{72.8} & 71.0 & \textbf{76.3} & \textbf{73.7} \\
\bottomrule
\end{tabular}}
\label{tab:kl_ablation}
\end{table*}

Recent works such as DAPO \cite{Yu2025DAPOAO, Vassoyan2025IgnoreTK} argue that KL regularization can unnecessarily constrain policy updates and recommend removing the KL penalty entirely to allow freer exploration. In contrast, \citet{Huang2024CorrectingTM} revisit divergence regularization and propose using a chi-squared penalty to better control overoptimization. Motivated by these findings, we ablate the effect of different divergence constraints in our reinforcement learning setup for spatial reasoning.

Table~\cref{tab:kl_ablation} reports results on CV-Bench 2D and 3D tasks \cite{Tong2024Cambrian1AF} for three variants of \textsc{SpatialThinker-3B}: (i) no KL penalty, (ii) chi-squared divergence penalty with a coefficient of 0.01, and (iii) our default KL divergence penalty with a coefficient of 0.01. Removing the KL penalty leads to a noticeable drop in performance, particularly on 3D tasks. Using a chi-squared divergence penalty underperforms both the no-penalty and KL variants on several subtasks, especially depth and distance reasoning. The KL-regularized model achieves the best overall performance, yielding a CV-Bench average of 73.7\% and providing the strongest results on 3D reasoning tasks.

These findings suggest that a modest KL penalty stabilizes policy updates and prevents reward overoptimization in our spatial reasoning setting, leading to more reliable improvements. While recent language-only alignment work has advocated for removing divergence constraints, our results indicate that retaining a small KL term remains beneficial for multimodal reasoning tasks where stability and coherent spatial grounding are crucial.



\begin{table}[t]
\caption{Results on abstract reasoning benchmarks. Lego Puzzles measure compositional reasoning over object arrangements, while BLINK Multi-View requires integrating multi-view spatial cues.}
\centering
\setlength{\tabcolsep}{6pt}
\resizebox{0.75\textwidth}{!}{%
\begin{tabular}{l|cc}
\toprule
\textbf{Model} & \textbf{Lego Puzzles} & \textbf{BLINK Multi-View} \\
\midrule
\rowpink \multicolumn{3}{c}{\textit{Proprietary and Open-Source MLLMs}} \\
GPT-4o-0513 & \textbf{57.7} & \textbf{54.1} \\
Claude-3.5-Sonnet-0620 & 53.6 & 51.9 \\
Qwen2.5-VL-3B & 29.9 & 42.9 \\
Qwen2.5-VL-7B & 35.8 & 44.4 \\
VLAA-Thinker-7B & 33.4 & 51.1 \\
SpaceThinker & 31.5 & 50.4 \\
SpaceOm & 32.0 & 48.9 \\
\midrule
\rowblue \multicolumn{3}{c}{\textit{Method Comparison (Trained on SpatialThinkerVQA)}} \\
Qwen2.5-VL-3B + SFT & 34.7 & 42.1 \\
Qwen2.5-VL-3B + Vanilla GRPO & 27.0 & 45.9 \\
\rowour \textbf{SpatialThinker-3B (Ours)} & 33.9 & 45.1 \\
Qwen2.5-VL-7B + SFT & 36.6 & 44.4 \\
Qwen2.5-VL-7B + Vanilla GRPO & 29.7 & 51.9 \\
\rowour \textbf{SpatialThinker-7B (Ours)} & \underline{37.7} & \underline{52.6} \\
\bottomrule
\end{tabular}}
\label{tab:abstract_reasoning_results}
\end{table}

\section{Additional Results: Abstract Reasoning}\label{sec:abs_reason}

To further evaluate the generalization capacity of \textsc{SpatialThinker}, we examine its performance on two abstract reasoning benchmarks: \textbf{Lego Puzzles} \cite{Tang2025LEGOPuzzlesHG}, which test compositional object reasoning and multi-step spatial reasoning, and \textbf{BLINK Multi-View} \cite{Fu2024BLINKML}, which requires integrating spatial cues across multiple viewpoints, including visual-spatial understanding and perspective understanding. These tasks are not part of the training distribution and measure the ability of models to extrapolate structured reasoning skills to abstract domains.

Across both tasks, \textsc{SpatialThinker-7B} achieves the highest open-source performance improving over generalist and spatial MLLMs, and scoring 37.7\% on Lego Puzzles and 52.6\% on BLINK Multi-View, closely approaching GPT-4o and surpassing Claude 3.5 Sonnet on the latter. Interestingly, we observe that vanilla GRPO provides competitive performance on BLINK Multi-View but underperforms on Lego Puzzles, suggesting that dense spatial rewards offer complementary signals that better support compositional reasoning. These results demonstrate that the spatial grounding learned through reinforcement learning transfers to more abstract domains that require compositional and multi-view integration skills.


\onecolumn 
\section{Detailed Results: CV-Bench}

\begin{table}[H]
\caption{Detailed breakdown of CV-Bench \cite{Tong2024Cambrian1AF} results across Count, Relation, Depth, and Distance subtasks.}
\centering
\setlength{\tabcolsep}{10pt} 
\resizebox{1\textwidth}{!}{%
\begin{tabular}{l|cccc|cc|c}
\toprule
\multirow{2}{*}{\textbf{Model}} & 
\multicolumn{4}{c|}{\textbf{CV-Bench Tasks}} & 
\multicolumn{2}{c|}{\textbf{CV-Bench}} & 
\multirow{2}{*}{\textbf{Avg.}} \\
 & Count & Relation & Depth & Distance & 2D & 3D & \\
\midrule
\rowgray \multicolumn{8}{c}{\textit{Proprietary Models}} \\
GPT-5-0807 & 68.3 & 94.5 & 92.2 & 88.6 & 81.4 & 90.4 & 85.9 \\
GPT-4o-0513 & 65.9 & 85.7 & 87.8 & 78.2 & 75.8 & 83.0 & 79.4 \\
Gemini-1.5-Pro & 70.4 & 85.2 & 82.4 & 72.8 & 77.8 & 77.6 & 77.7 \\
Claude-4-Sonnet-0514 & 63.3 & 83.3 & 83.0 & 85.4 & 73.3 & 84.2 & 78.8 \\
Claude 3.7 Sonnet & - & 74.2 & 85.8 & 84.2 & - & 85.0 & - \\
\midrule
\rowpink \multicolumn{8}{c}{\textit{Open-Source General MLLMs}} \\
Qwen2-VL-2B & 54.7 & 22.6 & 16.7 & 31.7 & 38.7 & 24.2 & 31.5 \\
Qwen2.5-VL-3B & 61.5 & 58.3 & 67.3 & 53.0 & 59.9 & 60.2 & 60.1 \\
Qwen2.5-VL-7B & 55.9 & 82.2 & 70.0 & 66.0 & 69.1 & 68.0 & 68.6 \\
Qwen3-VL-30B & 65.9 & 92.1 & 92.0 & 87.2 & 79.0 & 89.6 & 84.3 \\
VLAA-Thinker-3B & 61.6 & 83.5 & 53.0 & 46.8 & 72.6 & 49.9 & 61.3 \\
VLAA-Thinker-7B & 47.0 & 74.6 & 61.3 & 59.2 & 60.8 & 60.3 & 60.6 \\
LLaVA-NeXT-34B & - & - & - & - & 73.0 & 74.8 & 73.9 \\
Mini-Gemini-HD-34B & - & - & - & - & 71.5 & 79.2 & 75.4 \\
Cambrian-1-34B & - & - & - & - & 74.0 & 79.7 & 76.9 \\
\midrule
\rowgreen \multicolumn{8}{c}{\textit{Open-Source Spatial MLLMs}} \\
Spatial-LLaVA-7B & - & - & 57.3 & 52.2 & - & 54.8 & - \\
SATORI-R1 & 42.8 & 66.3 & 65.5 & 73.3 & 54.6 & 69.4 & 62 \\
VisualThinker-R1-2B & 59.6 & 66.8 & 54.2 & 56.7 & 63.2 & 55.45 & 59.3 \\
Spatial-RGPT-7B w/ depth & - & - & 62.3 & 59.0 & - & 60.7 & - \\
RoboPoint-13B & - & 75.6 & 77.8 & 44.5 & - & 61.15 & - \\
SpaceThinker-3B & 61.0 & 69.2 & 70.5 & 61.3 & 65.1 & 65.9 & 65.5 \\
SpaceLLaVA-13B & - & 63.7 & 66.8 & 70.2 & - & 68.5 & - \\
SpatialBot-3B & - & 69.4 & 77.3 & 60.8 & - & 69.05 & - \\
\midrule
\rowblue \multicolumn{8}{c}{\textit{Method Comparison (Trained on STVQA-7K)}} \\
Qwen2.5-VL-3B + SFT & 30.2 & 77.5 & 61.2 & 75.5 & 53.9 & 68.4 & 61.2 \\
Qwen2.5-VL-3B + Vanilla GRPO & 67.5 & 73.7 & 64.0 & 69.2 & 70.6 & 66.6 & 68.6 \\
\rowour \textbf{SpatialThinker-3B (Ours)} & 68.5 & 73.5 & 79.7 & 72.8 & 71.0 & 76.3 & 73.7 \\
Qwen2.5-VL-7B + SFT & 33.3 & 78.9 & 64.8 & 77.7 & 56.1 & 71.3 & 63.7 \\
Qwen2.5-VL-7B + Vanilla GRPO & 58.9 & 78.8 & 79.3 & 73.7 & 68.9 & 76.5 & 72.7 \\
\rowour \textbf{SpatialThinker-7B (Ours)} & 68.7 & 86.7 & 81.2 & 76.2 & 77.7 & 78.7 & 78.2 \\
\rowour \textbf{SpatialThinker-30B (Ours)} & 67.4 & 93.2 & 96.1 & 91.1 & 80.3 & 93.6 & 87.0 \\
\bottomrule
\end{tabular}}
\label{tab:cvbench_breakdown}
\end{table}


\clearpage
\onecolumn 
\section{Detailed Results: 3DSRBench}

\begin{table}[H]
\caption{Detailed Breakdown of 3DSRBench \cite{Ma20243DSRBenchAC} Height, Location, Orientation, and Multi-Object tasks.}
\centering
\setlength{\tabcolsep}{10pt} 
\resizebox{1\textwidth}{!}{%
\begin{tabular}{l|cccc|c}
\toprule
\multirow{2}{*}{\textbf{Model}} & 
\multicolumn{4}{c|}{\textbf{3DSRBench Tasks}} & 
\multirow{2}{*}{\textbf{Avg.}} \\
 & Height & Location & Orientation & Multi-Object & \\
\midrule
\rowgray \multicolumn{6}{c}{\textit{Proprietary Models}} \\
GPT-5-0807 & 72.9 & 79.5 & 59.0 & 60.6 & 68.2 \\
GPT-4o-0513 & 53.2 & 59.6 & 21.6 & 39.0 & 44.3 \\
Claude-4-Sonnet-0514 & 68.6 & 72.7 & 48.4 & 56.8 & 62.0 \\
Claude-3.5-Sonnet-0620 & 53.5 & 63.1 & 31.4 & 41.3 & 48.2 \\
Gemini 2.0 Flash & 49.7 & 68.9 & 32.2 & 41.5 & 49.9 \\
Gemini 2.0 Flash (thinking) & 53.0 & 67.1 & 35.8 & 43.6 & 51.1 \\
\midrule
\rowpink \multicolumn{6}{c}{\textit{Open-Source MLLMs}} \\
Qwen2.5-VL-3B & 45.2 & 56.8 & 35.7 & 35.7 & 44.0 \\
Qwen2.5-VL-7B & 44.1 & 62.7 & 40.6 & 40.5 & 48.4 \\
Qwen3-VL-30B & 59.1 & 75.4 & 51.0 & 54.6 & 60.4 \\
Qwen2.5-VL-72B & 53.3 & 71.0 & 43.1 & 46.6 & 54.9 \\
Cambrian-1-8B & 23.2 & 53.9 & 35.9 & 41.9 & 42.2 \\
LLaVA-NeXT-8B & 50.6 & 59.9 & 36.1 & 43.4 & 48.4 \\
VLAA-Thinker-7B & 54.0 & 60.2 & 42.9 & 49.1 & 52.2 \\
\midrule
\rowgreen \multicolumn{6}{c}{\textit{Open-Source Spatial MLLMs}} \\
SpatialBot-3B & 40.4 & 54.4 & 31.9 & 33.5 & 41.1 \\
SpaceLLaVA-13B & 49.3 & 54.4 & 27.6 & 35.4 & 42.0 \\
SpatialLLM-8B & 45.8 & 61.6 & 30.0 & 36.7 & 44.9 \\
SATORI-R1 & 50.9 & 52.3 & 41.9 & 46.3 & 48.0 \\
SpatialRGPT-7B w/ depth & 55.9 & 60.0 & 34.2 & 42.3 & 48.4 \\
SpaceThinker-3B & 53.1 & 57.3 & 41.9 & 49.6 & 51.1 \\
\midrule
\rowblue \multicolumn{6}{c}{\textit{Method Comparison (Trained on STVQA-7K)}} \\
Qwen2.5-VL-3B + SFT & 51.1 & 58.3 & 42.7 & 48.1 & 50.8 \\
Qwen2.5-VL-3B + Vanilla GRPO & 48.9 & 57.9 & 42.5 & 47.2 & 50.1 \\
\rowour \textbf{SpatialThinker-3B (Ours)} & 52.6 & 61.8 & 43.4 & 49.8 & 52.9 \\
Qwen2.5-VL-7B + SFT & 50.6 & 66.3 & 43.8 & 47.9 & 53.6 \\
Qwen2.5-VL-7B + Vanilla GRPO & 54.3 & 64.7 & 45.5 & 50.4 & 54.7 \\
\rowour \textbf{SpatialThinker-7B (Ours)} & 52.0 & 70.3 & 45.5 & 50.9 & 56.4 \\
\rowour \textbf{SpatialThinker-30B (Ours)} & 62.6 & 74.9 & 50.5 & 54.3 & 62.1 \\
\bottomrule
\end{tabular}}
\label{tab:height_location_orientation_multi}
\end{table}


\clearpage

\section{Detailed Results: STVQA-7K\textsubscript{val}}

\begin{table}[H]
\caption{Per-category accuracy on the held-out STVQA-7K\textsubscript{val} split across the nine spatial reasoning categories. \textsc{SpatialThinker} models lead across nearly all categories, with the largest gains on relational, size, and existence reasoning.}
\centering
\setlength{\tabcolsep}{4pt}
\resizebox{\textwidth}{!}{%
\begin{tabular}{l|ccccccccc|c}
\toprule
\textbf{Model} & \textbf{Relation} & \textbf{Reach} & \textbf{Size} & \textbf{Orient.} & \textbf{Location} & \textbf{Depth} & \textbf{Distance} & \textbf{Count} & \textbf{Existence} & \textbf{Overall} \\
\midrule
\rowgray \multicolumn{11}{c}{\textit{Proprietary Models}} \\
GPT-5-0807 & 89.4 & \underline{94.0} & 89.5 & 88.4 & 86.9 & 88.0 & \textbf{100.0} & 87.5 & 87.1 & 89.7 \\
GPT-4o-0513 & 75.8 & 82.1 & 64.7 & 84.7 & 78.4 & 72.1 & 90.1 & 84.5 & 80.8 & 77.0 \\
Claude-4-Sonnet-0514 & 79.2 & 78.0 & \underline{93.8} & 69.2 & 82.6 & 74.0 & 86.7 & 81.3 & 90.3 & 80.5 \\
\midrule
\rowpink \multicolumn{11}{c}{\textit{Open-Source General \& Spatial MLLMs}} \\
Qwen2.5-VL-3B & 72.6 & 80.0 & 79.2 & 73.1 & 78.3 & 72.0 & 73.4 & 78.1 & 74.2 & 74.3 \\
Qwen2.5-VL-7B & 77.0 & 80.2 & 85.6 & 81.0 & 80.6 & 78.2 & 73.5 & 65.8 & 74.4 & 77.5 \\
Qwen3-VL-30B & 84.5 & 86.0 & 81.3 & 80.8 & 82.6 & 80.0 & 83.4 & 87.5 & 80.7 & 83.7 \\
VLAA-Thinker-7B & 75.5 & 70.0 & 81.3 & 88.5 & 80.5 & 74.0 & 73.4 & 71.9 & 80.7 & 76.2 \\
SpaceThinker & 74.1 & 77.5 & 80.3 & 71.0 & 85.6 & 63.5 & 66.2 & 75.1 & \underline{96.6} & 75.4 \\
SpaceOm & 64.3 & 70.3 & 73.6 & 65.4 & 68.4 & 64.3 & 65.6 & 67.3 & 66.5 & 66.0 \\
SpatialReasoner & 75.9 & 72.9 & 72.1 & 81.4 & 66.8 & 72.9 & 66.3 & 71.1 & 72.0 & 74.0 \\
SATORI-R1 & 49.4 & 59.8 & 54.9 & 41.2 & 54.0 & 49.8 & 55.8 & 57.0 & 50.0 & 51.2 \\
Visionary-R1 & 73.6 & 80.0 & 77.1 & 73.1 & 78.2 & 72.0 & 66.6 & 78.1 & 74.2 & 74.4 \\
\midrule
\rowblue \multicolumn{11}{c}{\textit{Method Comparison (Trained on STVQA-7K)}} \\
Qwen2.5-VL-3B + SFT & 85.3 & 86.3 & 89.9 & 85.0 & 87.3 & 76.3 & 87.0 & 87.8 & 90.7 & 85.6 \\
Qwen2.5-VL-3B + Vanilla GRPO & 86.0 & 86.0 & 89.6 & 84.6 & 89.1 & 82.0 & 86.7 & 87.5 & \textbf{96.8} & 86.7 \\
\rowour \textbf{SpatialThinker-3B (Ours)} & 91.4 & 93.3 & \textbf{95.1} & \underline{95.4} & 97.1 & \textbf{93.3} & 92.6 & 86.8 & 96.1 & 92.5 \\
Qwen2.5-VL-7B + SFT & 84.5 & 80.1 & 85.5 & 88.6 & 84.9 & 78.1 & 86.8 & \textbf{93.9} & 84.0 & 84.5 \\
Qwen2.5-VL-7B + Vanilla GRPO & 87.3 & 82.0 & 87.5 & 80.7 & 93.4 & 82.0 & 86.6 & 90.6 & 93.5 & 87.1 \\
\rowour \textbf{SpatialThinker-7B (Ours)} & \underline{93.2} & \underline{94.0} & \underline{93.8} & 88.5 & \underline{97.9} & \underline{90.0} & 90.0 & 87.5 & 93.6 & \underline{92.8} \\
\rowour \textbf{SpatialThinker-30B (Ours)} & \textbf{93.4} & \textbf{94.1} & 93.4 & \textbf{98.5} & \textbf{99.2} & 80.1 & \underline{93.4} & \underline{91.3} & 93.9 & \textbf{93.0} \\
\bottomrule
\end{tabular}}
\label{tab:stvqa_breakdown}
\end{table}

\clearpage
\onecolumn 
\section{Additional Qualitative Results}\label{sec:add_model_eg}

\begin{figure}[H]
    \centering
    \includegraphics[width=0.8\linewidth]{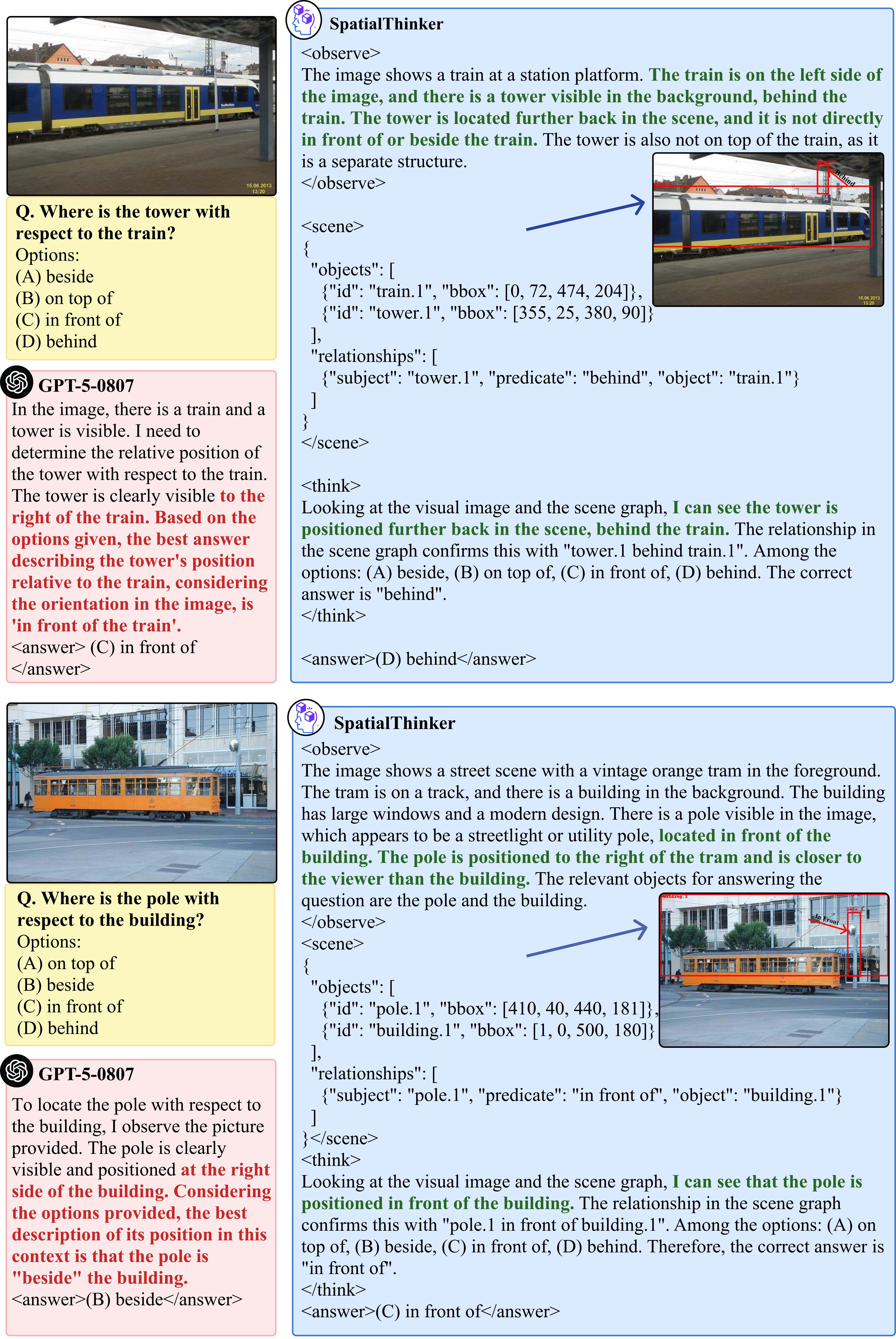}
    \caption{
        \textbf{Additional qualitative comparisons between GPT-5-0807 and SpatialThinker.}
        SpatialThinker shows stronger spatial grounding and fine-grained object distinction, accurately identifying 3D relations that GPT-5-0807 often confuses.
    }
    \label{fig:st-output-eg-more}
\end{figure}

\twocolumn






\end{document}